\documentclass[10pt,journal,compsoc]{IEEEtran}
\usepackage{lastpage} 
\usepackage{amssymb}
\usepackage{color}
\usepackage{amsmath}
\usepackage{amsthm}
\usepackage{amsmath}
\usepackage{amsfonts}
\usepackage{url}
\usepackage{multirow}
\usepackage{footnote}
\usepackage{amssymb}
\usepackage{mathtools}
\usepackage{algorithmicx}
\usepackage{algorithm}
\usepackage{algpseudocode}
\usepackage{cases}
\usepackage{cite}
\usepackage[caption=false,font=footnotesize]{subfig}
\usepackage{makecell}
\usepackage{tabularx}
\usepackage[table]{colortbl}
\usepackage{booktabs}
\usepackage{tikz}
\usetikzlibrary{arrows}
\usepackage{subfig}
\usepackage{dblfloatfix}
\usepackage{flushend}
\usepackage{standalone}
\usepackage{soul,xcolor}
\usepackage{color, colortbl}

\input{mysymbol.sty}
\usepackage{needspace}

\input{pennColors.sty}

\newcommand{\transp}{\mathsf{T}}

\definecolor{Gray}{gray}{0.75}

\setstcolor{black}

\ifCLASSINFOpdf
\else
\fi

\begin{document}

\title{EdgeNets: \!Edge Varying Graph Neural Networks}

\author{Elvin Isufi, Fer\hspace{0.15cm}nando Gama and Alejandro Ribeiro
\IEEEcompsocitemizethanks{\IEEEcompsocthanksitem Isufi is with the Intelligent Systems Department, Delft University of Technology, Delft, The Netherlands. Gama is with the Department of Electrical Engineering and Computer Sciences, University of California, Berkeley, CA 94709. Ribeiro is with the Department of Electrical and Systems Engineering, University of Pennsylvania, Philadelphia, PA 19104. Isufi conducted this research during his postdoctoral period at the Department of Electrical and Systems Engineering, University of Pennsylvania. This work is supported by NSF CCF 1717120, ARO W911NF1710438, ARL DCIST CRA W911NF-17-2-0181, ISTC-WAS and Intel DevCloud. E-mails: e.isufi-1@tudelft.nl, fgama@berkeley.edu, aribeiro@seas.upenn.edu. Part of this work has been presented in \cite{isufi2019generalizing}.
}
\thanks{}}

%
%


\IEEEtitleabstractindextext{%
\begin{abstract}
Driven by the outstanding performance of neural networks in the structured Euclidean domain, recent years have seen a surge of interest in developing neural networks for graphs and data supported on graphs. The graph is leveraged at each layer of the neural network as a parameterization to capture detail at the node level with a reduced number of parameters and computational complexity. Following this rationale, this paper puts forth a general framework that unifies state-of-the-art graph neural networks (GNNs) through the concept of EdgeNet. An EdgeNet is a GNN architecture that allows different nodes to use different parameters to weigh the information of different neighbors. By extrapolating this strategy to more iterations between neighboring nodes, the EdgeNet learns edge- and neighbor-dependent weights to capture local detail. {This is a general linear and local operation that a node can perform and encompasses under one formulation all existing graph convolutional neural networks (GCNNs) as well as graph attention networks (GATs).} In writing different GNN architectures with a common language, EdgeNets highlight specific architecture advantages and limitations, while providing guidelines to improve their capacity without compromising their local implementation. For instance, we show that GCNNs have a parameter sharing structure that induces permutation equivariance. This can be an advantage or a limitation, depending on the application. In cases where it is a limitation, we propose hybrid approaches and provide insights to develop several other solutions that promote parameter sharing without enforcing permutation equivariance. Another interesting conclusion is the unification of GCNNs and GATs ---approaches that have been so far perceived as separate. In particular, we show that GATs are GCNNs on a graph that is learned from the features. This particularization opens the doors to develop alternative attention mechanisms for improving discriminatory power.
%
\end{abstract}

}

\maketitle

\IEEEdisplaynontitleabstractindextext

\IEEEpeerreviewmaketitle

\IEEEraisesectionheading{\section{Introduction}}
\label{sec:intro}
\IEEEPARstart{D}{ata} generated by networks is increasingly common. Examples include user preferences in recommendation systems, writer proclivities in blog networks \cite{tang2009relational}, or properties of assembled molecular compounds \cite{wale2008comparison}. Different from data encountered in the structured temporal or spatial domains, network data lives in high-dimensional irregular spaces. This fact makes difficult to extend tools that exploit the regularity of time and space, leading to a rising interest in novel techniques for dealing with network data \cite{bronstein2017geometric}. Since graphs are the prominent mathematical tool to model individual node properties ---product ratings, writer bias, or molecule properties--- along with node dependencies ---user similarities, blog hyperlinks, or molecular bonds--- the interest in network data has translated into a concomitant increase in the interest in tools for processing graphs and data supported on graphs \cite{shuman2013emerging}.

Several recent works have proposed graph neural networks (GNNs) as a means of translating to graphs the success convolutional and recurrent neural networks have attained at learning on time and space \cite{scarselli2005graph,scarselli2008graph, Gama20-GNNs, bruna2013spectral, defferrard2016convolutional, gama2018convolutional, du2017topology, kipf2016semi, xu2018powerful, levie2017cayleynets, simonovsky2017dynamic, monti2017geometric, atwood2016diffusion, velickovic2017graph, wu2019comprehensive, zhou2018graph, zhang2018deep, lee2018attention}. GNNs are first concretized in \cite{scarselli2005graph,scarselli2008graph} by means of recursive neighboring label aggregations combined with pointwise nonlinearities. The convolutional GNN counterpart appears in \cite{bruna2013spectral} where graph convolutions are defined as pointwise operators in the Laplacian's spectrum.
To avoid the cost and numerical instability of spectral decompositions, \cite{defferrard2016convolutional} approximates this spectral convolution with a Chebyshev polynomial on the Laplacian matrix. Parallel to these efforts, the field of graph signal processing has developed notions of graph convolutional filters as polynomials on a matrix representation of a graph \cite{taubiny2000geometric,shuman2011distributed, sandryhaila2013discrete,narang2013compact,teke2016extending, segarra2017optimal, isufi2017autoregressive}. This has led to GNNs described as architectures that simply replace time convolutions with graph convolutions \cite{gama2018convolutional, du2017topology}. A third approach to define GNNs is to focus on the locality of convolutions by replacing the adjacency of points in time with the adjacency of neighbors in a graph; something that can be accomplished by mixing nodes' features with their neighbor's features \cite{kipf2016semi, xu2018powerful}.

Despite their different motivations, spectral GNNs \cite{bruna2013spectral, defferrard2016convolutional}, polynomial GNNs \cite{gama2018convolutional, du2017topology}, and  local GNNs \cite{kipf2016semi, xu2018powerful} can all be seen to be equivalent to each other (Section \ref{sec:gcnn}). {In particular, they all share the reuse of parameters across all neighborhoods of a graph as well as indifference towards the values of different neighbors --see also \cite{you2020graph}. This is an important limitation that is tackled, e.g., by the graph attention networks (GAT) of \cite{velickovic2017graph, wu2019comprehensive, zhou2018graph, zhang2018deep, lee2018attention} through the use of attention mechanisms \cite{bahdanau2014neural, vaswani2017attention}. In this paper, we leverage edge varying graph filter \cite{coutino2018advances} to provide a generic framework for the design of GNNs that can afford flexibility to use different parameters at different nodes as well as different weighing to different neighbors of a node (Section \ref{sec_gnn}).} Edge varying filters are linear finite order recursions that allow individual nodes to introduce weights that are specific to the node, specific to each neighbor, and specific to the recursion index. {In this way, the edge varying recursion represents a general linear operation that a node can implement locally. I.e., a general operation that relies on information exchanges only with neighbor nodes (Section \ref{sec_ev_graph_filters}).}

{In alternative to the EdgeNet, graph network \cite{battaglia2018relational} is a popular framework for generalizing GNNs. Graph network consist of general update and aggregation functions over nodal, edge, and entire graph features. This unification is slightly more general than the message passing neural network \cite{gilmer2017neural} and considers updates to be principally affected by information exchange only with the one-hop neighbors. While providing relevant insights on the local detail of order-one filter GNNs such as  \cite{kipf2016semi,velickovic2017graph}, this strategy does not put emphasis on the role of the filter within the GNN or how the parameters of such filter are allocated to the different multi-hop neighbours \cite{defferrard2016convolutional,gama2018convolutional}.  Instead, the EdgeNet framework focuses only on nodal feature aggregations to highlight the role of multi-hop exchanges within a layer and to put emphasis on how different solutions operate from a node perspective. At the same time, the EdgeNet framework allows for a filter spectral analysis \cite{ortega2018graph}, which provides a better understanding of the type of filters that conform the learned filter bank. This spectral perspective of GNNs will shed light, for instance, of the advantages of ARMA filters \cite{isufi2017autoregressive} in learning sharper responses with less parameters than finite impulse response graph filters \cite{shuman2011distributed,sandryhaila2013discrete,segarra2017optimal}.
}

In its most general form, the edge varying GNNs allocate different parameters to the different edges, which is of the order of the number of nodes and edges of the graph. While allocating different parameters over the edges can help exploiting the graph structure better for the learning task at hand, edge-specific parameters sacrifice the inductive capabilities across different graphs. To reduce the complexity of this parameterization, we can regularize EdgeNets in different ways by imposing restrictions on the freedom to choose different parameters at different nodes. We explain that existing GNN architectures are particular cases of EdgeNets associated with different parameter restrictions. In turn, this shows how these solutions sacrifice the degrees of freedom to gain in parameter sharing and inductive capabilities. We further utilize the insight of edge varying recursions to propose novel GNN architectures. In consequence, the novel contributions of this paper are:

\begin{list}{label}{\leftmargin = 19.7pt \labelwidth=19.7pt \topsep=3pt \itemsep=3pt}
\item [{\bf(i)}] We define EdgeNets, which parameterize the linear operation of neural networks through a bank of edge varying recursions. {EdgeNets are a generic framework to design GNN architectures (Section~\ref{sec_gnn}).}
\item [{\bf(ii)}] {We show the approaches in \cite{bruna2013spectral, defferrard2016convolutional, gama2018convolutional, du2017topology, kipf2016semi, xu2018powerful, levie2017cayleynets, simonovsky2017dynamic, monti2017geometric, atwood2016diffusion} (among others) are EdgeNets where all nodes share the parameters.} We extend the representing power of these networks by adding some level of variability in weighing different nodes and different edges of a node (Section \ref{sec:gcnn}).

\item [{\bf(iii)}] Replacing finite length polynomials by rational functions provides an alternative parameterization of convolutional GNNs in terms of autoregressive moving average (ARMA) convolutional graph filters \cite{isufi2017autoregressive}. These ARMA GNNs generalize rational functions based on Cayley polynomials \cite{levie2017cayleynets} (Section \ref{subsec:FIRGCNN}).

\item [{\bf(iv)}] {We show that GATs can
be understood as GNNs with convolutional graph filters where a graph is learned ad hoc in each layer to represent the required abstraction between nodes.} The weights of this graph choose neighbors whose values should most influence the computations at a particular node. This reinterpretation allows for the proposal of more generic GATs with higher expressive power (Section \ref{sec:gat}).

\end{list}

\noindent {The this paper has the following structure. Section \ref{sec_ev_graph_filters} reviews edge varying recursions on graphs and Section~\ref{sec_gnn} introduces edge varying GNNs. To ease exposition, the GNN are grouped into two categories: convolutional in Section~\ref{sec:gcnn} and attentional in Section~\ref{sec:gat}. Within each category, we follow the same rationale. First, we discuss the state-of-the-art solutions as a particular case of the EdgeNet framework. Then, we discuss their architectural advantages and limitations. Finally, we leverage the EdgeNet viewpoint to propose new solutions that address some of these limitations and highlight the corresponding tradeoffs. Section~\ref{sec:nr} evaluates these solutions with numerical results and Section~\ref{sec:conc} concludes the paper.
}

\section{Edge Varying Linear Graph Filters}
\label{sec_ev_graph_filters}

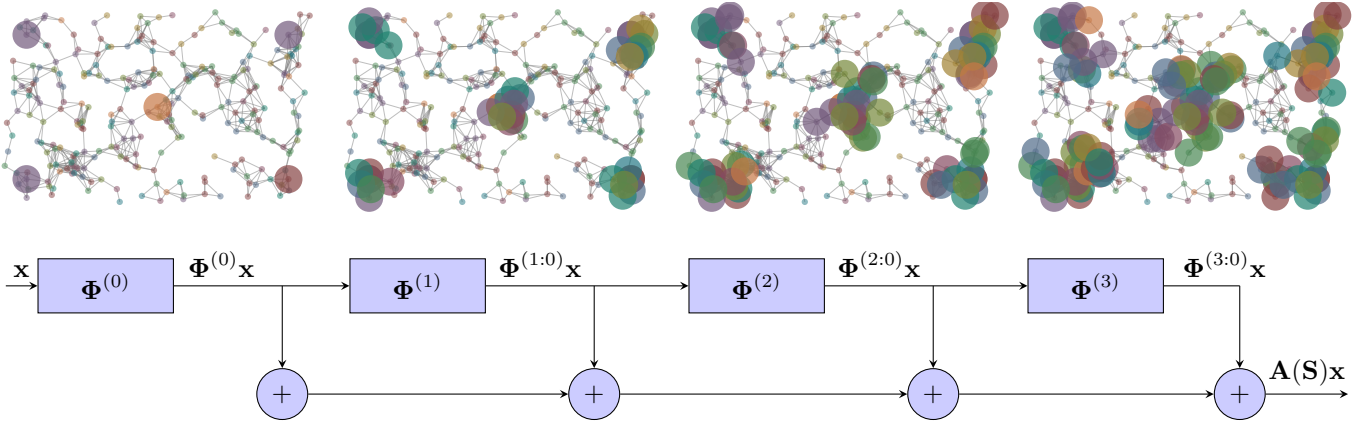
\begin{figure*}[t]
\centering
\includegraphics [width = 0.245\linewidth]
                 {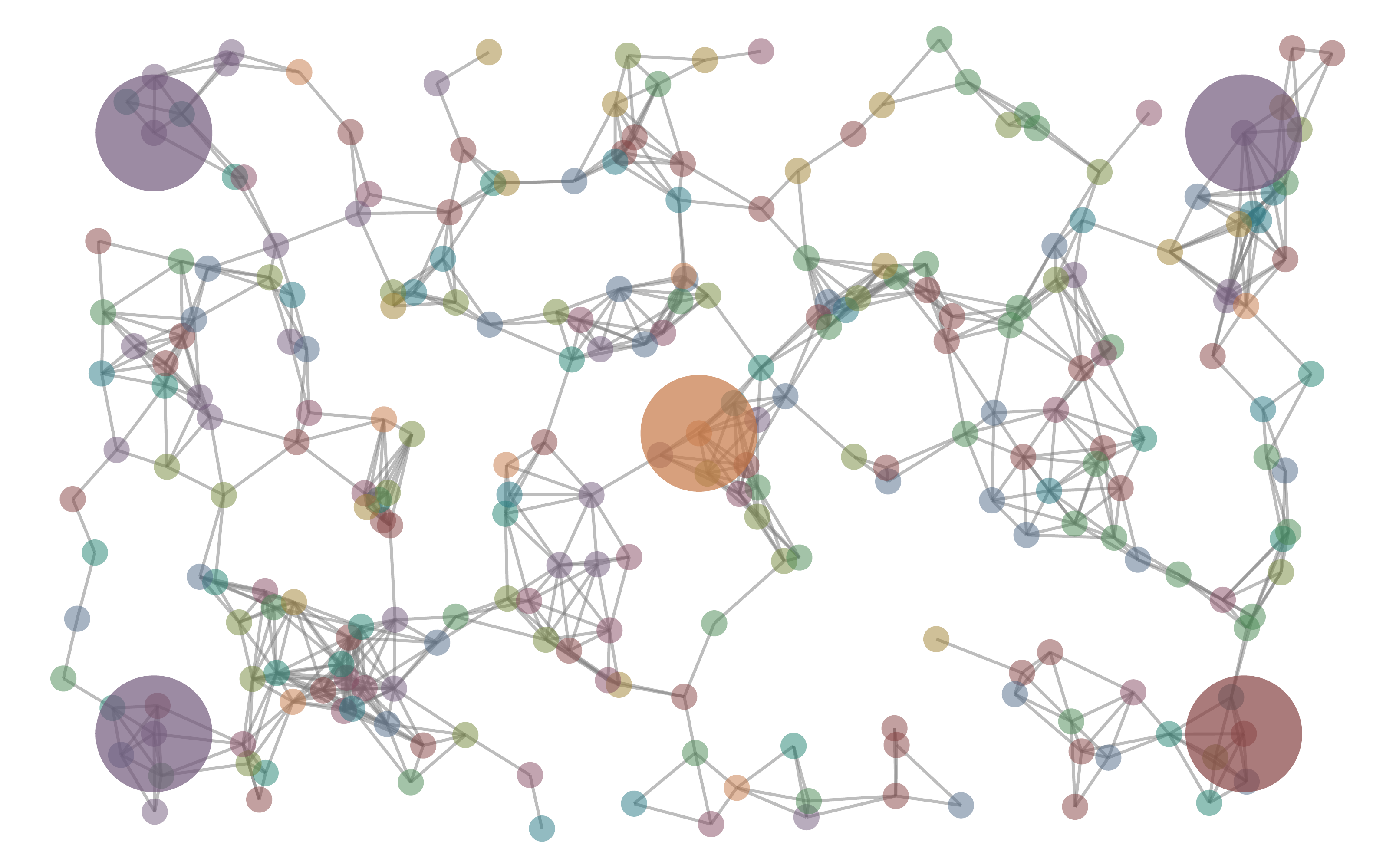}
\includegraphics [width = 0.245\linewidth]
                 {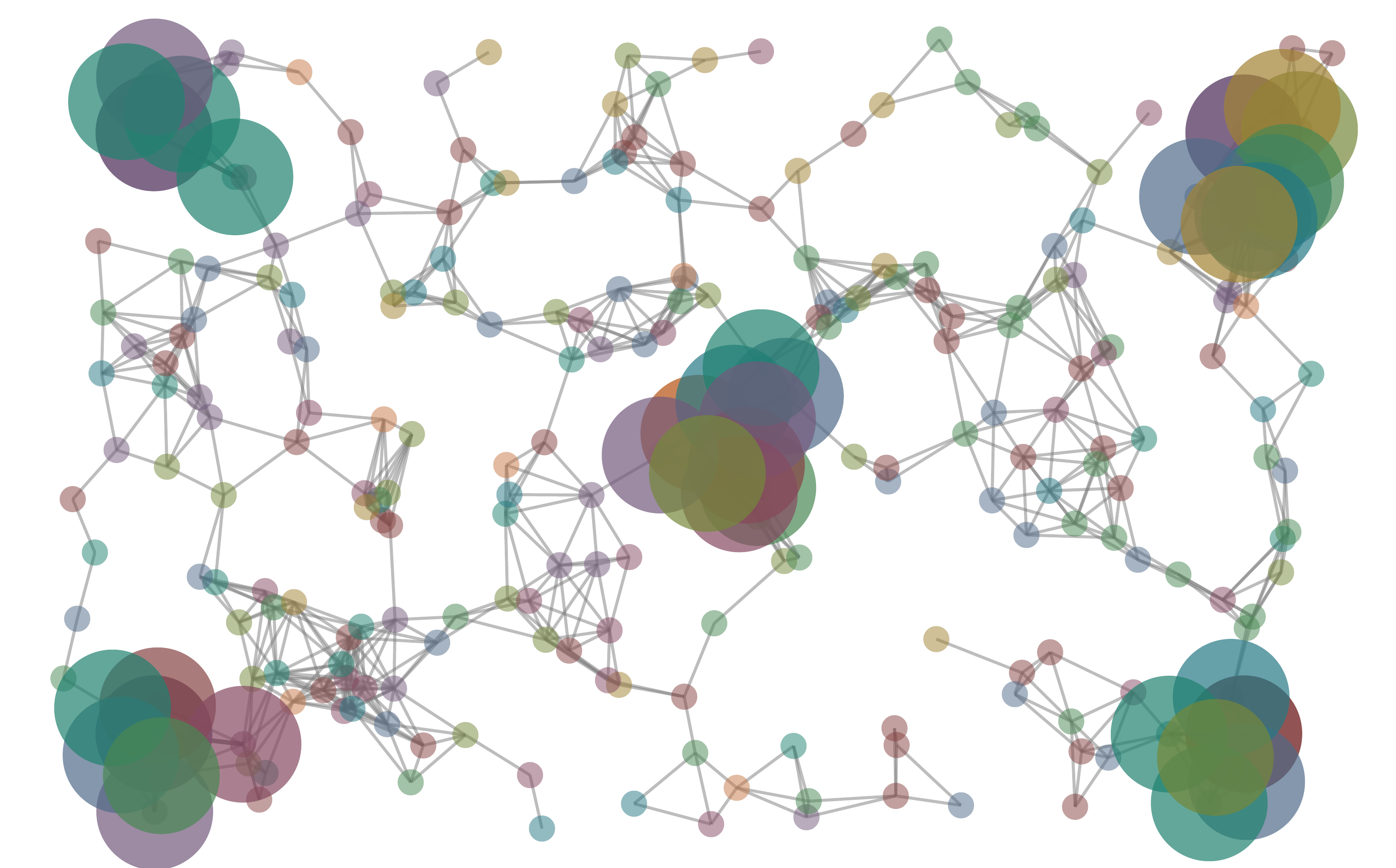}
\includegraphics [width = 0.245\linewidth]
                 {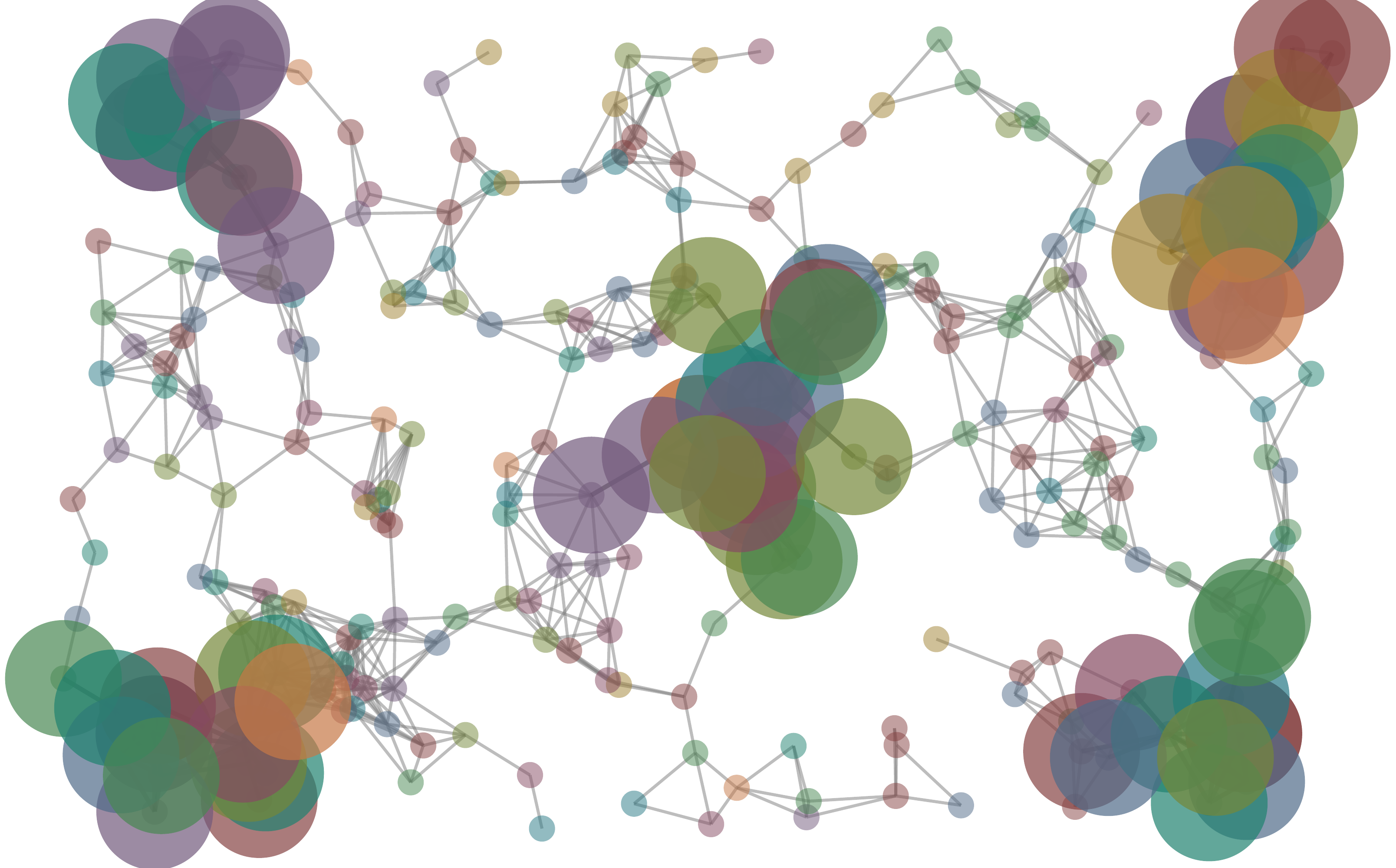}
\includegraphics [width = 0.245\linewidth]
                 {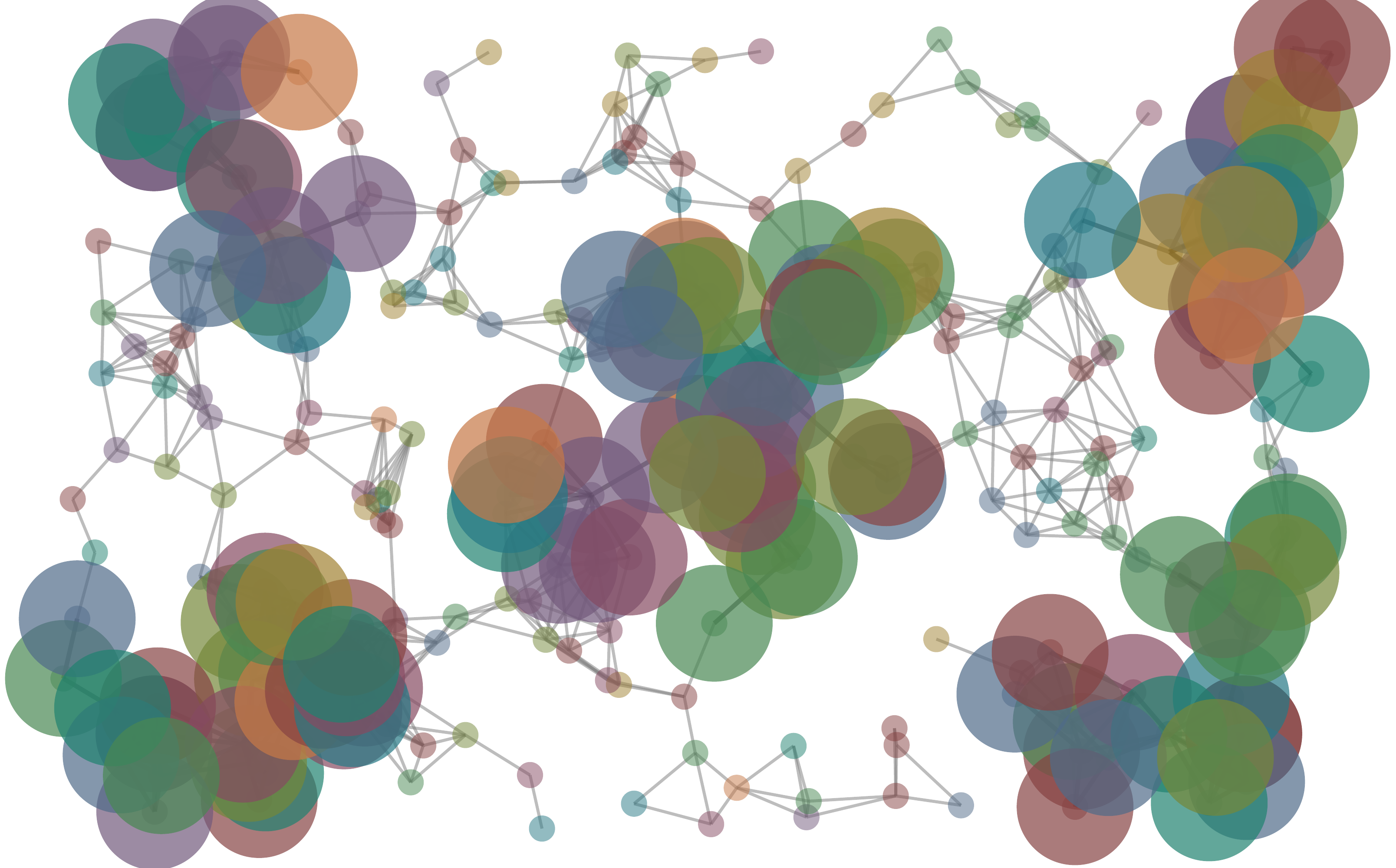} \\ \bigskip

\def \thisplotscale {1.8}
\def \unit {\thisplotscale cm}

\tikzstyle {Phi} = [rectangle, 
                    thin,
                    minimum width = 1.0*\unit, 
                    minimum height = \sumshift*\unit, 
                    anchor = west,
                    draw,
                    fill = blue!20]

\tikzstyle {sum} = [circle, 
                    thin,
                    minimum width  = 0.3*\unit, 
                    minimum height = 0.3*\unit, 
                    anchor = center,
                    draw,
                    fill = blue!20]

\def \deltax {2.0}
\def \deltay {0.8}
\def \sumshift {0.4}

\begin{tikzpicture}[x = 1*\unit, y = 1*\unit]
    
\node (first) [] {};    
    
\path (first) ++ (0.15*\deltax, 0) node (0) [Phi] {$\bbPhi^{(0)}$};
\path (0)     ++ (0.9*\deltax, 0) node (1) [Phi] {$\bbPhi^{(1)}$};
\path (1)     ++ (1.0*\deltax, 0) node (2) [Phi] {$\bbPhi^{(2)}$};
\path (2)     ++ (1.0*\deltax, 0) node (3) [Phi] {$\bbPhi^{(3)}$};

\path (3.east) ++ (0.7*\sumshift*\deltax, 0) node [anchor=west] (last) [] {};

\path[-stealth] (first) edge [above            ] node {$\bbx$}               (0);	
\path[-stealth] (0)     edge [above, near start] node {$\ \bbPhi^{(0)}\bbx$}   (1);	
\path[-stealth] (1)     edge [above, near start] node {$\ \bbPhi^{(1:0)}\bbx$} (2);	
\path[-stealth] (2)     edge [above, near start] node {$\ \bbPhi^{(2:0)}\bbx$} (3);\path[-]        (3)     edge [above, near end  ] node {$\ \bbPhi^{(3:0)}\bbx$} (last);				
\path (0.east) ++ (\sumshift*\deltax, -\deltay) node (sum0) [sum] {$+$};
\path (1.east) ++ (\sumshift*\deltax, -\deltay) node (sum1) [sum] {$+$};
\path (2.east) ++ (\sumshift*\deltax, -\deltay) node (sum2) [sum] {$+$};
\path (3.east) ++ (0.7*\sumshift*\deltax, -\deltay) node (sum3) [sum] {$+$};

\path[-stealth, draw] (sum0 |- 0) -- (sum0);	
\path[-stealth, draw] (sum1 |- 1) -- (sum1);	
\path[-stealth, draw] (sum2 |- 2) -- (sum2);	
\path[-stealth, draw] (sum3 |- 3) -- (sum3);	

\path[-stealth, draw] (sum0) -- (sum1);	
\path[-stealth, draw] (sum1) -- (sum2);	
\path[-stealth, draw] (sum2) -- (sum3);	

\path[-stealth] (sum3) edge [above] node 
                {$\bbA(\bbS)\bbx$} ++ (0.4*\deltax, 0);

\end{tikzpicture}
\caption{Edge Varying Graph Filters. Each edge varying matrix $\bbPhi^{(k)}$ acts as a different shift operator that locally combines the graph signal. (Top-left) The colored discs are centered at five reference nodes and their coverage shows the amount of local information needed to compute $\bbz^{(1)} = \bbPhi^{(1:0)}\bbx$ at these nodes. The coverage of the discs in the other graphs shows the signal information needed by the reference nodes to produce the successive outputs. (Bottom) Schematic illustration of the edge varying filter output of order $K = 3$.} 
\label{fig.evRecMain}
\end{figure*}

%
Consider a weighted graph $\ccalG$ with vertex set $\ccalV = \{1, \ldots, N\}$, edge set $\ccalE \subseteq \ccalV \times \ccalV$ composed of $|\ccalE|=M$ ordered pairs $(i,j)$, and weight function $\ccalW: \ccalE \to \mbR$. For each node $i$, define the neighborhood $\ccalN_i = \{j: (j,i)\in\ccalE\}$ as the set of nodes connected to $i$ and let $N_i := |\ccalN_i|$ denote the number of elements (neighbors) in this set. Associated with $\ccalG$ is a graph shift operator matrix $\bbS \in \reals^{N \times N}$ whose sparsity pattern matches that of the edge set, i.e., entry $S_{ij} \neq 0$ when $(j,i) \in \ccalE$ or when $i = j$. Supported on the vertex set are graph signals $\bbx = [x_1, \ldots, x_N]^\transp \in \reals^N$ in which component $x_i$ is associated with node $i \in \ccalV$. 

The adjacency of points in time signals or the adjacency of points in images codifies a sparse and local relationship between signal components. This sparsity and locality are leveraged by time or space filters. Similarly, $\bbS$ captures the sparsity and locality of the relationship between components of a signal $\bbx$ supported on $\ccalG$. It is then natural to take the shift operator as the basis for defining filters for graph signals. In this spirit, let $\bbPhi^{(0)}$ be an $N\times N$ diagonal matrix and $\bbPhi^{(1)}, \ldots, \bbPhi^{(K)}$ be a collection of $K$ matrices sharing the sparsity pattern of $\bbI_N + \bbS$. Consider then the sequence of signals $\bbz^{(k)}$ as
\begin{align}\label{eqn_z_nonrecursive}
   \bbz^{(k)} \ = \ \prod_{k'=0}^{k} \bbPhi^{(k')}  \bbx  
          \ = \ \bbPhi^{(k:0)}\bbx,  \quad
          \text{for\ } k=0,\ldots,K
\end{align}
where the product matrix $\bbPhi^{(k:0)}:=\prod_{k'=0}^{k} \bbPhi^{(k')} = \bbPhi^{(k)}\ldots\bbPhi^{(0)}$ is defined for future reference. Signal $\bbz^{(k)}$ can be computed using the recursion
\begin{align}\label{eqn_z_recursion}
   \bbz^{(k)} \ = \ \bbPhi^{(k)} \bbz^{(k-1)},  \quad
          \text{for\ } k= 0,\ldots,K
\end{align}
with initialization $\bbz^{(-1)}=\bbx$. This recursive expression implies signal $\bbz^{(k)}$ is produced from $\bbz^{(k-1)}$ using operations that are local in the graph. Indeed, since $\bbPhi^{(k)}$ shares the sparsity pattern of $\bbS$, node $i$ computes its component $z_{i}^{(k)}$ as
\begin{align}\label{eqn_z_recursion_local}
   z_{i}^{(k)} \ = \ \sum_{j\in \ccalN_i \cup i} \Phi^{(k)}_{ij}\, z_{j}^{(k-1)} .
\end{align}
Particularizing \eqref{eqn_z_recursion_local} to $k=0$, it follows each node $i$ builds the $i$th entry of $\bbz^{(0)}$ as a scaled version of its signal $\bbx$ by the diagonal matrix $\bbPhi^{(0)}$, i.e., $z_i^{(0)} = \Phi_{ii}^{(0)}x_i$. Particularizing to $k=1$, \eqref{eqn_z_recursion_local} yields the components of $\bbz^{(1)}$ depend on the values of the signal $\bbx$ at most at neighboring nodes. Particularizing to $k=2$, \eqref{eqn_z_recursion_local} shows the components of $\bbz^{(2)}$ depend only on the values of signal $\bbz^{(1)}$ at neighboring nodes which, in turn, depend only on the values of $\bbx$ at their neighbors. Thus, the components of $\bbz^{(2)}$ are a function of the values of $\bbx$ at most at the respective two-hop neighbors. Repeating this argument iteratively, $z_{i}^{(k)}$ represents an aggregation of information at node $i$ coming from its $k$-hop neighborhood ---see Figure \ref{fig.evRecMain}.

The collection of signals $\bbz^{(k)}$ behaves like a sequence of scaled shift operations except that instead of shifting the signal in time, the signal is diffused through the graph (the signal values are \emph{shifted} between neighboring nodes). Leveraging this interpretation, the graph filter output $\bbu$ is defined as the sum
\begin{align}\label{eqn_convolution_def}
   \bbu \ = \ \sum_{k=0}^K \bbz^{(k)}
        \ = \ \sum_{k=0}^K \bbPhi^{(k:0)}\, \bbx .
\end{align}
A filter output in time is a sum of scaled and shifted copies of the input signal. That \eqref{eqn_convolution_def} behaves as a filter follows from interpreting $\bbPhi^{(k:0)}$ as a scaled shift, which holds because of its locality. Each shift $\bbPhi^{(k:0)}$ is a recursive composition of individual shifts $\bbPhi^{(k)}$. These individual shifts represent different operators that respect the structure of $\ccalG$ while reweighing individual edges differently when needed. 

For future reference, define the filter matrix $\bbA(\bbS)$ so \eqref{eqn_convolution_def} rewrites as $\bbu = \bbA(\bbS)\bbx$. For this to hold, the filter matrix must be
\begin{align}\label{eqn_evFG}
   \bbA(\bbS)
        \ = \ \sum_{k=0}^K \bbPhi^{(k:0)}
        \ = \ \sum_{k=0}^K \Bigg(\prod_{k'=0}^{k} \bbPhi^{(k')}\Bigg).
\end{align}
{Following \cite{coutino2018advances}, $\bbA(\bbS)$ is a $K$th order edge varying graph filter. }Each matrix $\bbPhi^{(k)}$ contains at most $M+N$ nonzero elements corresponding to the nonzero entries of $\bbI_N + \bbS$; thus, the total number of parameters defining filter $\bbA(\bbS)$ in \eqref{eqn_evFG} is $K(M+N) + N$. For short filters, this is smaller than the $N^2$ components of an arbitrary linear transform. Likewise, in computing $\bbz^{(k)} =\bbPhi^{(k)} \bbz^{(k-1)}$ as per \eqref{eqn_z_recursion} incurs a computational complexity of order $\ccalO(M+N)$. This further results in an overall computational complexity of order $\ccalO\big(K(M+N)\big)$ for obtaining the filter output $\bbu$ in \eqref{eqn_convolution_def}. This reduced number of parameters and computational cost is leveraged next to define graph neural network (GNN) architectures with a controlled number of parameters and computational complexity matched to the graph sparsity.

\begin{remark} \normalfont
{The presence of the edge $(j,i)$ in graph $\ccalG$ is interpreted here as signal components $x_j$ and $x_i$ being related by the given structure in the data.} The shift operator entry $S_{ij}$ is a measure of the expected similarity. Larger entries indicate linked signal components are more related to each other. Therefore, the definition of the shift operator $\bbS$ makes it a valid stand-in for any graph representation matrix. Forthcoming discussions are valid whether $\bbS$ is an adjacency or a Laplacian matrix in any of their various normalized and unnormalized forms. We use $\bbS$ to keep discussions generic. \qed
\end{remark}


\section{Edge Varying Graph Neural Networks}
\label{sec_gnn}


Edge varying graph filters are the basis for defining GNN architectures through composition with pointwise nonlinear functions. Formally, consider a set of $L$ layers indexed by $l=1,\ldots,L$ and let $\bbA_l(\bbS) =  \sum_{k=0}^K \bbPhi_l^{(k:0)}$ be the graph filter used at layer $l$. A GNN is defined by the recursive expression
\begin{align}\label{eqn_gnn_single_feature}
   \bbx_l \ = \ \sigma \Big( \bbA_l(\bbS)\, \bbx_{l-1} \Big)
          \ = \ \sigma \Bigg( \sum_{k=0}^K \bbPhi_{l}^{(k:0)}\, \bbx_{l-1} \Bigg)
\end{align}
where we convene that $\bbx_0=\bbx$ is the input to the GNN and $\bbx_L$ is its output. To augment the representation power of GNNs, it is customary to add multiple node features per layer. We do this by defining matrices $\bbX_l = [\bbx_{l}^1, \ldots, \bbx_{l}^{F_l}]\in\reals^{N\times F_l}$ in which each column $\bbx_{l}^f$ represents a different graph signal at layer $l$. These so-called features are cascaded through layers where they are processed with edge varying graph filters and composed with pointwise nonlinearities according to
\begin{align}\label{eqn_gnn_multiple_feature}
   \bbX_l \ = \ \sigma \Bigg( \sum_{k=0}^K \bbPhi_{l}^{(k:0)}\, \bbX_{l-1} \bbA_{lk} \Bigg)
\end{align}
where $\bbA_{lk}\in\reals^{F_{l-1} \times F_{l}}$ is a parameter matrix that affords flexibility to process different features with different filter parameters. It is ready to see that \eqref{eqn_gnn_multiple_feature} represents a bank of edge varying graph filters $\bbA_{l}^{fg}(\bbS)$ applied to a set of $F_{l-1}$ input features $\bbx_{l-1}^g$ to produce a set of $F_l$ output features $\bbx_{l}^f$. Indeed, if we let $a_{lk}^{fg} = [\bbA_{lk}]_{fg}$ denote the $(f,g)$th entry of $\bbA_{lk}$, \eqref{eqn_gnn_multiple_feature} produces a total of $F_{l-1} F_l$ intermediate features of the form\footnote{{Throughout the paper, we will denote \emph{any} graph filter by $\bbA_{l}^{fg}(\bbS)$ to indicate that it is a matrix depending on the graph shift operator $\bbS$. When this filter contains additional parameters rather than those on the edges (i.e., $\bbPhi_l^{fg,(k:0)}$), we will indicate them with scalars $a_{lk}^{fg}$ [cf. \eqref{eqn_gnn_intermediate_features}]. For consistency, when expressing the bank of filters in a single recursion [cf. \eqref{eqn_gnn_multiple_feature}], we will group parameters $a_{lk}^{fg}$ into the matrix $\bbA_{lk}$.	}}
\begin{align}\label{eqn_gnn_intermediate_features}
   \bbu_{l}^{fg} \ = \ \bbA_{l}^{fg}(\bbS)\, \bbx_{l-1}^{g}
                 \ = \ \sum_{k=0}^K  a_{lk}^{fg}\, \bbPhi_{l}^{fg,(k:0)}\, \bbx_{l-1}^{g} 
\end{align}
for $g=1,\ldots,F_{l-1}$ and $f=1,\ldots,F_{l}$. The features $\bbu_{l}^{fg}$ are then aggregated across all $g$ and passed through a pointwise nonlinearity to produce the output features of layer $l$ as
\begin{align}\label{eqn_gnn_output_features}
   \bbx_{l}^{f} = \sigma\Bigg( \, \sum_{g=1}^{F_{l-1}}\bbu_{l}^{fg} \, \Bigg).
\end{align}
At layer $l=1$ the input feature is a graph signal $\bbx_{0}^{1} = \bbx$. This feature is passed through $F_1$ filters to produce $F_1$ higher-level features as per \eqref{eqn_gnn_intermediate_features}. The latter are then processed by a pointwise nonlinearity [cf. \eqref{eqn_gnn_output_features}] to produce $F_1$ output features $\bbx^{f}_{1}$. The subsequent layers $l>1$ start with $F_{l-1}$ input features $\bbx_{l-1}^{g}$ that are passed through the filter bank $\bbA_{l}^{fg}(\bbS)$ [cf. \eqref{eqn_gnn_intermediate_features}] to produce the higher-level features $\bbu_{l}^{fg}$. These are aggregated across all $g=1,\ldots,F_{l-1}$ and passed through a nonlinearity to produce the layer's output features $\bbx_{l}^{f}$ [cf. \eqref{eqn_gnn_output_features}]. In the last layer $l=L$, we consider without loss of generality the number of output features is $F_L =1$. This single feature $\bbx^{1}_L = \bbx_{L}$ is the output of the edge varying GNN or, for short, EdgeNet. {Remark the EdgeNet aggregates at each layer information form neighbors that are up to $K$ hops away [cf. \eqref{eqn_gnn_multiple_feature}]. This increases its flexibility to process intermediate features and generalizes the masking aggregation rule in \cite{yan2018spatial}, which can be seen as an EdgeNet of order $K = 1$.}

The EdgeNet output is a function of the input signal $\bbx$ and the collection of filter banks $\bbA^{fg}_{l}$ [cf. \eqref{eqn_evFG}]. {Group the filters in the filter tensor $\ccalA(\bbS)=\{\bbA_{l}^{fg}(\bbS)\}_{lfg}$ so that to define the GNN output as the mapping}
\begin{align}\label{eqn_gnn}
    \bbPsi \Big( \bbx; \ccalA(\bbS) \Big) := \bbx_L  \quad
   \text{with\ }
   \ccalA(\bbS) = \Big\{\bbA_{l}^{fg}(\bbS)\Big\}_{lfg}.
\end{align}
The filter parameters are trained to minimize a loss over a training set of input-output pairs $\ccalT= \{(\bbx, \bby)\}$. This loss measures the difference between the EdgeNet output $\bbx_{L}$ and the true value $\bby$ averaged over the examples $(\bbx, \bby)\in\ccalT$. 

As it follows from \eqref{eqn_evFG}, the number of parameters in each filter is $K(M+N)+N$. This gets scaled by the number of filters per layer $F_{l-1} F_{l}$ and the number of layers $L$. To provide an order bound on the number of parameters defining the EdgeNet set the maximum feature number $F=\max_{l} F_l$ and observe the number of parameters per layer is of order $(K(M+N) +N)F^2+F^2$. Likewise, the computational complexity at each layer is of order $\ccalO\big(K(M+N)F^{2}\big)$. This number of parameters and computational complexity are expected to be smaller than the corresponding numbers of a fully connected neural network. This is a consequence of exploiting the sparse nature of edge varying filters [cf. \eqref{eqn_convolution_def} and \eqref{eqn_evFG}]. A GNN can be then considered as an architecture that exploits the graph structure to reduce the number of parameters of a fully connected neural network. The implicit hypothesis is those signal components associated with different nodes are processed together in accordance with the nodes' proximity in the graph. 

We will show different existing GNN architectures are particular cases of \eqref{eqn_gnn_intermediate_features}-\eqref{eqn_gnn_output_features} using different subclasses of edge varying graph filters (Section \ref{sec:gcnn}) and the same is true for graph attention networks (Section \ref{sec:gat}). Establishing these relationships allows the proposal of natural architectural generalizations that increase the descriptive power of GNNs while still retaining manageable complexity. 


{\begin{remark}\label{rem.limit} \normalfont
The key property of the EdgeNet is to allocate trainable prarameters for each edge in each shift. While this formulation improves the expressive power of a GNN, it affects its inductive capability over graphs (i.e., the ability to generalize to new unseen graphs) \cite{hamilton2017inductive}. In the full form \eqref{eqn_gnn_multiple_feature}, the EdgeNet does not have inductive capabilities for graphs but only for graph signals. That is, it cannot be applied to test cases where new unseen graphs are present but can only be applied to test cases where new unseen graph signals are present for a fixed graph support. We shall see in the next section that graph convolutional neural networks sacrifice instead the degrees of freedom to gain inductive capabilities also for graphs.
\end{remark}
}

\begin{remark}\label{rem.multiEfeat} \normalfont
In the proposed EdgeNet, we considered graphs with single edge features, i.e., each edge is described by a single scalar. However, even when the graph has multiple edge features, say $E$, the EdgeNet extends readily to this scenario. This can be obtained by seeing the multi-edge featured graph as the union of $E$ graphs $\ccalG_{e}=(\ccalV, \ccalE_{e})$ with identical node set $\ccalV$ and respective shift operator matrix $\bbS_e$. For $\{\bbPhi^{e(k)}\}$ being the collection of the edge varying parameter matrices [cf. \eqref{eqn_z_nonrecursive}] relative to the shift operator $\bbS_e$, the $l$th layer output $\bbX_{l}$ [cf. \eqref{eqn_gnn_multiple_feature}] becomes
\begin{equation} \label{eqn_gnn_multiple_edge}
	\bbX_{l} = \sigma \left(
		\sum_{e=1}^{E}
			\sum_{k=0}^{K} \bbPhi_{l}^{e(k:0)} \bbX_{l-1} \bbA_{lk}^{e}
	\right).
\end{equation}
I.e., the outputs of each filter are aggregated also over the edge-feature dimension. The number of parameters and computational complexity get scaled by $E$. The GNN architectures discussed in the remainder of this manuscript, as a special case of the EdgeNet, are readily extendable to the multi-edge feature scenario by replacing \eqref{eqn_gnn_multiple_feature} with \eqref{eqn_gnn_multiple_edge}. The approach in \cite{ioannidis2019recurrent} is the particular case for \eqref{eqn_gnn_multiple_edge} with $K\!\! = \!\!1$ and the parameter matrix reduced to a scalar.
\end{remark}


\section{Graph Convolutional Neural Networks}
\label{sec:gcnn}


\def \bbphi {\bba}

{Graph convolutional neural networks (GCNNs) have shown great success to learning representations for graph data with prominent variants introduced in \cite{defferrard2016convolutional, kipf2016semi, du2017topology, gama2018convolutional}.
}
%
All these variants be written as GNN architectures in which the edge varying component in \eqref{eqn_gnn_multiple_feature} is fixed and given by powers of the shift operator matrix $\bbPhi_{l}^{(k:0)} = \bbS^{k}$, 
\begin{equation}\label{eqn_gcnn_matrix_notation}
   \bbX_l = \sigma \Bigg( \sum_{k=0}^K  \bbS^{k} \bbX_{l-1} \bbA_{lk}\Bigg) .
\end{equation}
{By comparing \eqref{eqn_gnn_intermediate_features} with \eqref{eqn_gcnn_matrix_notation}, it follows this particular restriction yields a tensor $\ccalA(\bbS)$ with filters of the form}
%
\begin{equation}\label{eqn_gcnn_features}
   \bbA_{l}^{fg}(\bbS) \ = \ \sum_{k=0}^K a_{lk}^{fg}\, \bbS^{k}
\end{equation}
for some order $K$ and scalar parameters $ a_{l0}^{fg} , \ldots,  a_{lK}^{fg}$. Our focus in this section is to discuss variations on \eqref{eqn_gcnn_features}. To simplify the discussion, we omit the layer and feature indices and for the remainder of this section write 
\begin{equation}\label{eqn_gcnn}
   \bbA(\bbS) \ = \ \sum_{k=0}^K a_{k} \bbS^{k}.
\end{equation}
The filters in \eqref{eqn_gcnn} are of the form in \eqref{eqn_evFG} with $\bbPhi^{(0)} = a_0 \bbI_N$ and $\bbPhi^{(k:0)} = a_k\bbS^k$ for $k \ge 1$. By particularizing $\ccalG$ to the line graph, \eqref{eqn_gcnn} represents a linear time-invariant filter described by a regular convolution. This justifies using the qualifier convolutional for an architecture with filters of the form \eqref{eqn_gcnn}. 

The appeal of the graph convolutional filters in \eqref{eqn_gcnn} is that they reduce the number of parameters from the $K(M+N) + N$ of the edge varying filters in \eqref{eqn_evFG} to just $K+1$; yielding also a computational complexity of order $\ccalO(KM)$. While we can reduce the number of parameters in several ways, the formulation in \eqref{eqn_gcnn} is of note because it endows the resulting GNN with equivariance to permutations of the labels of the graph. We state this property formally in the following proposition.

%
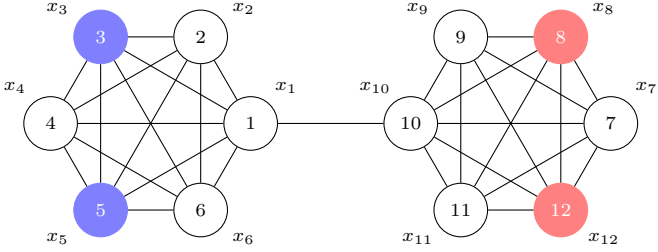
\begin{figure}[t!]
\centering
{\fontsize{7}{7}\selectfont\begin{tikzpicture}[scale = \myfactor]

  \node                         []     (center) {}; 
  \path (center) ++ (  0:\radius) node [empty node] (1) {$1$}  ++ ( 0.5,  0.5) node {$x_1$};
  \path (center) ++ ( 60:\radius) node [empty node] (2) {$2$}  ++ ( 0.6,  0.4) node {$x_2$};
  \path (center) ++ (120:\radius) node [blue node ] (3) {$3$}  ++ (-0.6,  0.4) node {$x_3$};
  \path (center) ++ (180:\radius) node [empty node] (4) {$4$}  ++ (-0.5,  0.5) node {$x_4$};
  \path (center) ++ (240:\radius) node [blue node ] (5) {$5$}  ++ (-0.6, -0.4) node {$x_5$};
  \path (center) ++ (300:\radius) node [empty node] (6) {$6$}  ++ ( 0.6, -0.4) node {$x_6$};

  \path (center) ++ (3.6*\radius,0) node [] (center) {}; 
  \path (center) ++ (  0:\radius) node [empty node] (7)  {$7$}   ++ ( 0.5,  0.5) node {$x_7$};
  \path (center) ++ ( 60:\radius) node [red node  ] (8)  {$8$}   ++ ( 0.6,  0.4) node {$x_8$};
  \path (center) ++ (120:\radius) node [empty node] (9)  {$9$}   ++ (-0.6,  0.4) node {$x_9$};
  \path (center) ++ (180:\radius) node [empty node] (10) {$10$}  ++ (-0.5,  0.5) node {$x_{10}$};
  \path (center) ++ (240:\radius) node [empty node] (11) {$11$}  ++ (-0.6, -0.4) node {$x_{11}$};
  \path (center) ++ (300:\radius) node [red node  ] (12) {$12$}  ++ ( 0.6, -0.4) node {$x_{12}$};

  \path (1)  edge [edge] node {} (2);
  \path (2)  edge [edge] node {} (3);
  \path (3)  edge [edge] node {} (4);
  \path (4)  edge [edge] node {} (5);
  \path (5)  edge [edge] node {} (6);
  \path (6)  edge [edge] node {} (1);
  \path (1)  edge [edge] node {} (3);
  \path (2)  edge [edge] node {} (4);
  \path (3)  edge [edge] node {} (5);
  \path (4)  edge [edge] node {} (6);
  \path (5)  edge [edge] node {} (1);
  \path (6)  edge [edge] node {} (2);
  \path (1)  edge [edge] node {} (4);
  \path (2)  edge [edge] node {} (5);
  \path (3)  edge [edge] node {} (6);

  \path (7)   edge [edge] node {}  (8);
  \path (8)   edge [edge] node {}  (9);
  \path (9)   edge [edge] node {} (10);
  \path (10)  edge [edge] node {} (11);
  \path (11)  edge [edge] node {} (12);
  \path (12)  edge [edge] node {}  (7);
  \path (7)   edge [edge] node {}  (9);
  \path (8)   edge [edge] node {} (10);
  \path (9)   edge [edge] node {} (11);
  \path (10)  edge [edge] node {} (12);
  \path (11)  edge [edge] node {}  (7);
  \path (12)  edge [edge] node {}  (8);
  \path (7)   edge [edge] node {} (10);
  \path (8)   edge [edge] node {} (11);
  \path (9)   edge [edge] node {} (12);

  \path (1)  edge [edge] node {}  (10);

\end{tikzpicture}} 
\caption{Permutation equivariance of machine learning on graphs. Many tasks in machine learning on graphs are equivariant to permutations (cf. Proposition \ref{prop:invariance}) but not all are. E.g., we expect agents 3, 5, 8, and 12 to be interchangeable from the perspective of predicting product ratings from the ratings of other nodes. But from the perspective of community classification we expect 3 and 5 or 8 and 12 to be interchangeable, but 3 and 5 are not interchangeable with 8 and 12.}
\label{fig.permInv}
\end{figure}

%
\begin{proposition}\label{prop:invariance}
Let $\bbx$ be a graph signal defined on the vertices of a graph $\ccalG = (\ccalV, \ccalE)$ with shift operator $\bbS$. Consider also the output of a GCNN $\bbPsi(\bbx; \ccalA(\bbS))$ [cf. \eqref{eqn_gnn}] with input $\bbx$ and tensor $\ccalA(\bbS) = \big\{\bbA(\bbS)\big\}$ composed of filters of the form in \eqref{eqn_gcnn}. Then, for a permutation matrix $\bbP$, it holds that
\begin{equation*}
\bbP^\transp \bbPsi(\bbx; \ccalA(\bbS)) = \bbPsi(\bbP^\transp\bbx; \ccalA(\bbP^\transp\bbS\bbP)).
\end{equation*}
That is, the GCNN output operating on the graph $\ccalG$ with input $\bbx$ is a permuted version of the GCNN output operating on the permuted graph $\ccalG^\prime = (\ccalV^\prime, \ccalE^\prime)$ with permuted shift operator $\bbS^\prime = \bbP^\transp\bbS\bbP$ and permuted input signal $\bbx^\prime = \bbP^\transp\bbx$.
\end{proposition}

%
\begin{proof} See Appendix~\ref{sec:proofP1}.  \end{proof}

%
Proposition \ref{prop:invariance} establishes the output of a GCNN is independent of node labeling. This is important not just because graph signals are independent of labeling ---therefore, so should be their processing--- but because it explains how GCNNs exploit the internal signal symmetries. {If two parts of the graph are topologically identical and the nodes support identical signal values, a GCNN yields identical outputs \cite{maron2018invariant,levie2019transferability,gama2019stability}.} 

It must be emphasized that permutation equivariance is of use only inasmuch as this is a desirable property of the considered task. Permutation equivariance holds in, e.g., recommendation systems but does not hold in, e.g., community classification. In the graph in Figure~\ref{fig.permInv}, we expect agents 3, 5, 8, and 12 to be interchangeable from the perspective of predicting product ratings from the ratings of other nodes. But from the perspective of community classification, we expect 3 and 5 or 8 and 12 to be interchangeable, but 3 and 5 are not interchangeable with 8 and 12.

When equivariance is not a property of the task, GCNNs are not expected to do well. GCNNs will suffer in any problem in which local detail around a node is important. This is because the filter in \eqref{eqn_gcnn} forces all nodes to weigh the information of all $k$-hop neighbors with the same parameter $a_k$ irrespectively of the relative importance of different nodes and different edges. To avoid this limitation, we can use a GNN that relies on the edge varying filters \eqref{eqn_evFG} in which each node $i$ learns a different parameter $\Phi_{ij}^{(k)}$ for each neighbor $j$. {These two cases are analogous to CNNs processing time signals with conventional convolutional filters as opposed to a neural network that operates with arbitrarily time varying filters (i.e., filters whose coefficients change in time)}. The appealing intermediate solution is to use filters with {\it controlled edge variability} to mix the advantage of a permutation equivariant parameterization (Proposition~\ref{prop:invariance}) with the processing of local detail. We introduce architectures that construct different versions of filters with controlled edge variability in Sections \ref{subsec:nodeDep}-\ref{subsec:FIRGCNN}.

%
\begin{remark}\normalfont Along with the above-referred works, also the works in \cite{simonovsky2017dynamic,monti2017geometric,atwood2016diffusion} and \cite{xu2018powerful} use versions of the convolutional filter in \eqref{eqn_gcnn}. In specific, \cite{simonovsky2017dynamic,monti2017geometric, atwood2016diffusion} consider single shifts on the graph with shift operator a learnable weight matrix, a Gaussian kernel, and a random-walk, respectively. The work in \cite{xu2018powerful} adopts multi-layer perceptrons along the feature dimension at each node, before exchanging information with their neighbors. This is equivalent to \eqref{eqn_gcnn_matrix_notation} with the first layers having order $K=0$ (depending on the depth of the MLP), followed by a final layer of order $K=1$.
\end{remark}

%
\subsection{GNNs with Controlled Edge Variability}\label{subsec:nodeDep}

To build a GNN that fits between a permutation equivariant GCNN [cf. \eqref{eqn_gcnn}] and a full edge varying GNN [cf. \eqref{eqn_evFG}], we use different filter parameters in different parts of the graph. Formally, let $\ccalB=\{\ccalB_1,\ldots, \ccalB_B\}$ be a partition of the node set into $B$ blocks with block $\ccalB_i$ having $B_i$ nodes. Define the tall matrix $\bbC_\ccalB \in \{0,1\}^{N \times B}$ such that $[\bbC_{\ccalB}]_{ij} = 1$ if node $i$ belongs to block $\ccalB_{j}$ and $0$ otherwise. Let also $\bbphi^{(k)}_{\ccalB}\in\reals^B$ be a vector of block parameters of filter order $k$. {\it Block varying} graph filters are then defined as
\begin{equation}\label{eq.nvGCNN}
   \bbA(\bbS) = \sum_{k = 0}^K\diag\left(\bbC_{\ccalB}\bbphi^{(k)}_{\ccalB}\right)\bbS^k.
\end{equation}
Filters in \eqref{eq.nvGCNN} use parameters $[\bbphi^{(k)}_{\ccalB}]_i$ for all nodes $i\in \ccalB_i$.

Block varying filters belong to the family of node varying graph filters \cite{segarra2017optimal} and are of the form in \eqref{eqn_evFG} with
\begin{equation}\label{eqn_block_filters}
   \bbPhi^{(k:0)} = \diag(\bbC_{\ccalB}\bbphi^{(k)}_{\ccalB})\bbS^k.
\end{equation}
Substituting \eqref{eqn_block_filters} into \eqref{eqn_evFG} generates block varying GNNs \cite{gama2018convolutionalNV}. Block varying GNNs have $B(K+1)F^2$ parameters per layer and a computational complexity of order $\ccalO(KF^2M)$.

\begin{figure}[t!]
\centering
 
{\fontsize{7}{7}\selectfont \begin{tikzpicture}[scale = \myfactor]
  
  \node                         [empty node] (1)  {$1$};
  \path (1) + ( 2.3, 1.25) node [blue node ] (2)  {$2$};
  \path (1) + ( 2.3,-1.25) node [empty node] (3)  {$3$};
  \path (2) + ( 2.3, 0   ) node [empty node] (4)  {$4$}; 
  \path (3) + ( 2.3, 0   ) node [empty node] (5)  {$5$};
  \path (4) + ( 2.3, 0   ) node [empty node] (6)  {$6$}; 
  \path (5) + ( 2.3, 0   ) node [red node]   (7)  {$7$};
  \path (6) + ( 2.2,-1.25) node [empty node] (8)  {$8$};

  \path (1) edge [blue edge]  node [above left]  {$[\Phi^{(k)}_{\ccalI}]_{21}$} (2);
  \path (1) edge [light edge] node [above]       {}                             (3);
  \path (2) edge [light edge] node [above]       {}                             (1);
  \path (2) edge [light edge] node [above]       {}                             (3);
  \path (2) edge [light edge] node [above]       {}                             (4);
  \path (2) edge [light edge] node [right]       {}                             (5);
  \path (3) edge [light edge] node [below]       {}                             (1);
  \path (3) edge [blue edge]  node [left]        {$[\Phi^{(k)}_{\ccalI}]_{31}$} (2);
  \path (3) edge [light edge] node [left ]       {}                             (4);
  \path (3) edge [light edge] node [left ]       {}                             (5);
  \path (4) edge [blue edge]  node [below]       {$[\Phi^{(k)}_{\ccalI}]_{41}$} (2);
  \path (4) edge [light edge] node [above]       {}                             (3);
  \path (4) edge [light edge] node [above]       {}                             (6);
  \path (4) edge [red edge]   node [left]        {$[\Phi^{(k)}_{\ccalI}]_{74}$} (7);
  \path (5) edge [blue edge]  node [right]       {$[\Phi^{(k)}_{\ccalI}]_{51}$} (2);
  \path (5) edge [light edge] node [below]       {}                             (3);
  \path (5) edge [light edge] node [left ]       {}                             (6);
  \path (5) edge [red edge]   node [below]       {$[\Phi^{(k)}_{\ccalI}]_{75}$} (7);
  \path (6) edge [light edge] node [right]       {}                             (4);
  \path (6) edge [light edge] node [right]       {}                             (5);
  \path (6) edge [red edge]   node [right]       {$[\Phi^{(k)}_{\ccalI}]_{76}$} (7);
  \path (6) edge [light edge] node [above]       {}                             (8);
  \path (7) edge [light edge] node [below]       {}                             (4);
  \path (7) edge [light edge] node [below]       {}                             (5);
  \path (7) edge [light edge] node [below]       {}                             (6);
  \path (7) edge [light edge] node [below]       {}                             (8);
  \path (8) edge [light edge] node [below]       {}                             (6);
  \path (8) edge [red edge]   node [below right] {$[\Phi^{(k)}_{\ccalI}]_{78}$} (7);

  
\end{tikzpicture}} 
\caption{Hybrid Edge Varying Filter [cf. \eqref{eq.nvGCNN1}]. The nodes in set $\ccalI = \{2, 7\}$ are highlighted. Nodes $2$ and $7$ have edge varying parameters associated with their incident edges. All nodes, including 2 and 7, also use the global parameter $a_k$ as in a regular convolutional graph filter.}
\label{fig.nvEVRec}
\end{figure}
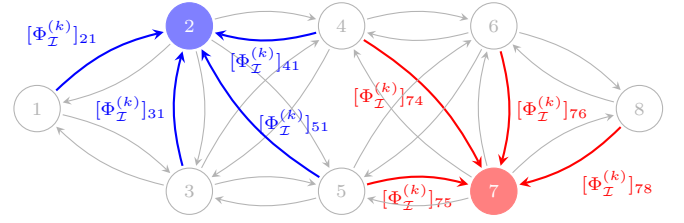

Alternatively, we can consider what we call {\it hybrid} filters that are defined as linear combinations of convolutional filters and edge varying filters that operate in a subset of nodes ---see Figure~\ref{fig.nvEVRec}. Formally, let $\ccalI\subset\ccalV$ denote an important subset of $I=|\ccalI|$ nodes and define the shift matrices $\bbPhi^{(k)}_{\ccalI}$ such that the diagonal matrix $\bbPhi_\ccalI^{(0)}$ has entries $[\bbPhi^{(0)}_{\ccalI}]_{ii} \neq 0$ for all $i \in \ccalI$ and $[\bbPhi^{(k)}_{\ccalI}]_{ij}=0$ for all $i\notin\ccalI$ or $(i,j)\notin\ccalE$ and $k \ge 1$. That is, the parameter matrices $\bbPhi^{(k)}_{\ccalI}$ may contain nonzero elements only at rows $i$ that belong to set $\ccalI$ and with the node $j$ being a neighbor of $i$. We define hybrid filters as those of the form
\begin{equation}\label{eq.nvGCNN1}
   \bbA(\bbS) = \sum_{k = 0}^K\bigg( \prod_{k^\prime = 0}^k\bbPhi_\ccalI^{(k^\prime)} 
                      + a_k\bbS^k\bigg).
\end{equation}
Substituting \eqref{eq.nvGCNN1} in \eqref{eqn_evFG} generates hybrid GNNs. In essence, nodes $i\in\ccalI$ learn edge dependent parameters which may also be different at different nodes, while nodes $i\notin\ccalI$ learn global parameters.

Hybrid filters are defined by a number of parameters that depends on the total neighbors of all nodes in the importance set $\ccalI$. Define then $M_\ccalI=\sum_{i\in\ccalI} N_i$ and observe $\bbPhi^{(0)}_{\ccalI}$ has $I$ nonzero entries since it is a diagonal matrix, while $\bbPhi^{(k)}_{\ccalI}$ for $k\geq 1$ have respectively $M_\ccalI$ nonzero values. We then have $KM_\ccalI+I$ parameters in the edge varying filters and $K+1$ parameters in the convolutional filters. We therefore have a total of $(I +  KM_\ccalI + K + 1)F^2$ parameters per layer in a hybrid GNN. The implementation cost of a hybrid GNN layer is of order $\ccalO(KF^2(M + N))$ since both terms in \eqref{eq.nvGCNN1} respect the graph sparsity. 

Block GNNs depend on the choice of blocks $\ccalB$ and hybrid GNNs on the choice of the importance set $\ccalI$. We explore the use of different heuristics based on centrality and clustering measures in Section~\ref{sec:nr} where we will see that the choice of $\ccalB$ and $\ccalI$ is in general problem specific.

%

%
\subsection{Spectral Graph Convolutional Neural Networks}\label{subsec:spGCNN}

The convolutional operation of the graph filter in \eqref{eqn_gcnn} can be represented in the spectral domain. To do so, consider the input-output relationship $\bbu = \bbA(\bbS)\bbx$ along with the eigenvector decomposition of the shift operator $\bbS = \bbV\bbLambda\bbV^{-1}$. Projecting the input and output signals in the eigenvector space of $\bbS$ creates the so-called graph Fourier transforms $\tbx := \bbV^{-1}\bbx $ and $\tbu := \bbV^{-1}\bbu$ \cite{sandryhaila2014discrete}\, which allow us to write 
\begin{equation}\label{eqn_polResp}
   \tbu  :=  \bigg( \sum_{k=0}^K a_k \bbLam^{k}\bigg) \tbx .
\end{equation}
Eq. \eqref{eqn_polResp} reveals convolutional graph filters are pointwise in the spectral domain, due to the diagonal nature of the eigenvalue matrix $\bbLambda$. We can, therefore, define the filter's spectral response $a : \reals \to \reals$ as the function
\begin{equation}\label{eqn_freq_response}
   a(\lam) = \sum_{k = 0}^K a_k\lam^k
\end{equation}
which is a single-variable polynomial characterizing the graph filter $\bbA(\bbS)$. If we allow for filters of order $K=N-1$, there is always a set of parameters $a_k$ such that $a(\lam_i)=\tda_i$ for any set of spectral response $\tda_i$ \cite{sandryhaila2013discrete}. Thus, training over the set of spectral parameters $a(\lam_1),\ldots, a(\lam_N)$ is equivalent to training over the space of (nodal) parameters $a_0, \ldots, a_{N-1}$. {GCNNs were first introduced in \cite{bruna2013spectral} using the spectral representation of graph filters in \eqref{eqn_freq_response}.}

By using edge varying graph filters [cf. \eqref{eqn_evFG}], we can propose an alternative parameterization of the space of filters of order $N$ which we will see may have some advantages. To explain this better let $\ccalJ$ be the index set defining the zero entries of $\bbS + \bbI_N$ and let $\bbC_\ccalJ \in \{0,1\}^{|\ccalJ|\times N^2}$ be a binary selection matrix whose rows are those of $\bbI_{N^2}$ indexed by $\ccalJ$. Let also $\bbB$ be a basis matrix that spans the null space of 
\begin{equation}\label{eqn.nulSpace}
\bbC_\ccalJ\text{vec}(\bbV^{-1}*\bbV)
\end{equation}
where $\text{vec}(\cdot)$ is the column-wise vectorization operator and $``*"$ is the Khatri-Rao product \cite{liu2008hadamard}. Then, the following proposition from \cite{coutino2018advances} quantifies the spectral response of a particular class of the edge varying graph filter in \eqref{eqn_evFG}.

\begin{proposition}\label{prop:EVResp} Consider the subclass of the edge varying graph filters in \eqref{eqn_evFG} where the parameter matrices $\big[\bbPhi^{(0)} + \bbPhi^{(1)}\big]$ and $\bbPhi^{(k)}$ for all $k = 2, \ldots, K$ are restricted to the ones that share the eigenvectors with $\bbS$, i.e., $\big[\bbPhi^{(0)} + \bbPhi^{(1)}\big] = \bbV\bbLambda^{(1)}\bbV^{-1}$ and $\bbPhi^{(k)} = \bbV\bbLambda^{(k)}\bbV^{-1}$ for all $k = 2, \ldots, K$. The spectral response of this subclass of edge varying filter has the form
\begin{align}\label{eq.EV_respfinProp}
\begin{split}
a(\bbLambda) = \sum_{k = 1}^K\bigg(\prod_{k^\prime = 1}^k\bbLambda^{(k^\prime)}	\bigg) =\sum_{k=1}^{K}\prod_{k^\prime = 1}^k\diag\left(\bbB\bbmu^{(k^\prime)}\right)
\end{split}
\end{align}
where $\bbB$ is an $N \times b$ basis kernel matrix that spans the null space of \eqref{eqn.nulSpace} and $\bbmu^{(k)}$ is a $b \times 1$ vector containing the expansion parameters of $\bbLambda^{(k)}$ into $\bbB$.
\end{proposition}

\begin{proof} See Appendix~\ref{sec:AppPropEV}.  \end{proof}

Proposition~\ref{prop:EVResp} provides a subclass of the edge varying graph filters where, instead of learning $K(M+N)+N$ parameters, they learn the $Kb$ entries $\bbmu^{(1)}, \ldots, \bbmu^{(K)}$ in \eqref{eq.EV_respfinProp}. These filters build the output features as a pointwise multiplication between the filter spectral response $a(\bbLambda)$ and the input spectral transform $\tbx = \bbV^{-1}\bbx$, i.e., $\bbu = \bbV a(\bbLambda)\tbx = \bbV a(\bbLambda)\bbV^{-1}\bbx$. Following then the analogies with conventional signal processing, \eqref{eq.EV_respfinProp} represents the spectral response of a convolutional edge varying graph filter. Spectral GCNNs are a particular case of \eqref{eq.EV_respfinProp} with order $K = 1$ and kernel $\bbB$ independent from the graph (e.g., a spline kernel). {Besides generalizing \cite{bruna2013spectral}, a graph-dependent kernel allows to implement \eqref{eq.EV_respfinProp} in the vertex domain through an edge varying filter of the form \eqref{eqn_evFG}; hence, having a complexity of order $\ccalO(K(M+N))$ in contrast to $\ccalO(N^2)$ required for the graph-independent kernels.} The edge varying implementation captures also local detail up to a region of radius $K$ from a node; yet, having a spectral interpretation. Nevertheless, both the graph-dependent GNN [cf. \eqref{eq.EV_respfinProp}] and the graph-independent GNN \cite{bruna2013spectral} are more of theoretical interest since they require the eigendecomposition of the shift operator $\bbS$. This aspect inadvertently implies a cubic complexity in the number of nodes and an accurate learning process will suffer from numerical instabilities since it requires an order $K \approx N$; hence, high order matrix powers $\bbS^k$.

%


%
\subsection{ARMA graph convolutional neural networks}\label{subsec:FIRGCNN}

We can increase the descriptive power of the filter in \eqref{eqn_gcnn} by growing its order $K$, which allows learning filters with a more discriminative polynomial frequency response [cf. \eqref{eqn_freq_response}]. However, this also increases the parameters and computational cost. Most importantly, it introduces numerical issues associated with high order matrix powers $\bbS^k$, ultimately, leading to poor interpolatory and extrapolatory performance \cite{heckert2003nist}. These challenges can be overcame by considering graph filters with a rational spectral response, since rational functions have better interpolatory and extrapolatory properties than polynomials \cite{heckert2003nist,trefethen2019approximation,liu2018filter}. Rational functions can also achieve more complicated responses with lower degrees in both numerator and denominator, thus, having less learnable parameters. Autoregressive moving average (ARMA) graph filters \cite{isufi2017autoregressive} serve for such purpose and implement rational functions of the form
%
%
%
\begin{equation}\label{eqn_arma}
   \bbA(\bbS) = \bigg({\bbI + \sum_{p = 1}^P a_{p}\bbS^p}	\bigg)^{-1}
                      {\bigg(\sum_{q = 0}^Q b_{q}\bbS^q\bigg)}
              := \bbP^{-1}(\bbS) \bbQ(\bbS)
\end{equation}
where we have defined $\bbP(\bbS):={\bbI + \sum_{p = 1}^P a_{p}\bbS^p}$ and $\bbQ(\bbS):=\sum_{q = 0}^Q b_{q}\bbS^q$. The ARMA filter in \eqref{eqn_arma} is defined by $P$ denominator parameters $\bba=[a_1, \ldots, a_{P}]^\top$ and $Q+1$ numerator parameters $\bbb=[b_0, \ldots, b_Q]^\top$. The input-output relationship $\bbu = \bbA(\bbS)\bbx$ of the ARMA filter can be represented in the spectral domain as [cf. \eqref{eqn_polResp}]
\begin{equation}\label{eqn_arma_spectral}
   \tbu = \bigg({\bbI + \sum_{p = 1}^P a_{p}\bbLam^p}	\bigg)^{-1}
                      \bigg(\sum_{q = 0}^Q b_{q}\bbLam^q\bigg) \, \tbx.
\end{equation}
It follows that ARMA filters are also pointwise operators in the spectral domain characterized by the rational spectral response function
\begin{equation}\label{eqn_arma_frequency_response}
   a(\lam) = \Big(\sum_{q = 0}^Q b_{q}\lam^q\Big) \ \Big/  \ \Big({1 + \sum_{p = 1}^P a_{p}\lam^p} \Big) .
\end{equation}
In particular, it follows the space of ARMA filters defined by \eqref{eqn_arma} is equivalent to the space of spectral ARMA filters defined by \eqref{eqn_arma_frequency_response} which is equivalent to the space of spectral filters in \eqref{eqn_freq_response} and, in turn, equivalent to the graph convolutional filters in \eqref{eqn_gcnn}. That they are equivalent does not mean they have the same properties. We expect ARMA filters produce useful spectral responses with less parameters than the convolutional filters in \eqref{eqn_gcnn} or the spectral filters in \eqref{eqn_freq_response}.

As it follows from \eqref{eqn_arma}, we need to compute the inverse matrix $\bbP^{-1}(\bbS)$ to get the ARMA output. The latter incurs a cubic complexity, which unless the graph is of limited dimensions is computationally unaffordable. When the graph is large, we need an iterative method that exploits the sparsity of the graph to approximate the inverse with a reduced cost \cite{isufi2017autoregressive, liu2018filter}. Due to its faster convergence, we consider a parallel structure that consists of first transforming the polynomial ratio in (18) in its partial fraction decomposition form and subsequently using the Jacobi method to approximate inverse. While also other Krylov approaches are possible, the parallel Jacobi method offers a better tradeoff between computational complexity and convergence rate.

\medskip\noindent{\bf Partial fraction decomposition of ARMA filters.} The partial fraction decomposition of the rational function $a(\lam)$ in \eqref{eqn_arma_frequency_response} provides an equivalent representation of ARMA filters. Let $\bbgamma=[\gamma_1, \ldots, \gamma_P]^\top$ be a set of poles, $\bbbeta=[\beta_1, \ldots, \beta_P]^\top$ a corresponding set of residuals and $\bbalpha=[\alpha_0, \ldots, \alpha_K]^\top$ be a set of direct terms; we can then rewrite \eqref{eqn_arma_frequency_response}  as
\begin{equation}\label{eqn_arma_partial}
   a(\lam) = \sum_{p = 1}^P \frac{\beta_p}{\lam - \gamma_p}
                     + \sum_{k = 0}^{K} \alpha_k\lambda^k
\end{equation}
where $\bbalpha$, $\bbbeta$, and $\bbgamma$ are computed from $\bba$ and $\bbb$. A graph filter whose spectral response is as in \eqref{eqn_arma_partial} is one in which the spectral variable $\lam$ is replaced by the shift operator variable $\bbS$. It follows that if $\bbalpha$, $\bbbeta$, and $\bbgamma$ are chosen to make \eqref{eqn_arma_partial} and \eqref{eqn_arma_frequency_response} equivalent, the filter in \eqref{eqn_arma} is, in turn, equivalent to
\begin{equation}\label{eqn_arma_partial_out}
   \bbA(\bbS) =  \sum_{p = 1}^P \beta_p \Big(\bbS - \gamma_p\bbI\Big)^{-1}
                          + \sum_{k = 0}^{K} \alpha_k\bbS^k .
\end{equation}
The equivalence of \eqref{eqn_arma} and \eqref{eqn_arma_partial_out} means that instead of training $\bba$ and $\bbb$ in  \eqref{eqn_arma} we can train $\bbalpha$, $\bbbeta$, and $\bbgamma$ in \eqref{eqn_arma_partial_out}. 

\medskip\noindent{\bf Jacobi implementation of single-pole filters.} To circumvent the matrix inverses in \eqref{eqn_arma_partial_out}, we first consider each single-pole filter in \eqref{eqn_arma_partial_out} separately and implement the input-output relationship
%
%
%
\begin{equation}\label{eqn_FIR_direct}
   \bbu_p  =  \beta_p \Big(\bbS - \gamma_p\bbI\Big)^{-1} \bbx.
\end{equation}
Expression \eqref{eqn_FIR_direct} is equivalent to the linear equation $(\bbS - \gamma_p\bbI)\bbu_p = \beta_p\bbx$, which we can solve iteratively through a Jacobi recursion. This requires us to separate $(\bbS - \gamma_p\bbI)$ into diagonal and off-diagonal components. We, therefore, begin by defining the diagonal degree matrix $\bbD=\diag(\bbS)$ so that the shift operator can be written as
\begin{equation}\label{eqn_degree_and_non_degree}
   \bbS \  = \ \bbD + \big(\bbS-\bbD\big)
        \ := \ \diag(\bbS) + \big(\bbS -\diag(\bbS)\big).
\end{equation}
With this definition, we write $(\bbS - \gamma_p\bbI_N) = (\bbD - \gamma_p\bbI_N) + (\bbS-\bbD\big)$, which is a decomposition on diagonal terms $(\bbD - \gamma_p\bbI_N)$ and off-diagonal terms $(\bbS-\bbD\big)$. The Jacobi iteration $k$ for \eqref{eqn_FIR_direct} is given by the recursive expression
\begin{equation}\label{eqn_jacobi_recursion}
   \bbu_{pk}  =  -\Big(\bbD - \gamma_p\bbI_N\Big)^{-1} \,
                        \Big[ \beta_p\bbx -\Big(\bbS-\bbD\Big)\bbu_{p(k-1)}\Big]
\end{equation}
initialized with $\bbu_{p0}=\bbx$. We can unroll this iteration to write an explicit relationship between $\bbu_{pk}$ and $\bbx$. To do that, we define the parameterized shift operator
\begin{equation}\label{eqn_jacobi_shift}
   \bbR (\gamma_p)  =  -\Big(\bbD - \gamma_p\bbI_N\Big)^{-1} \Big (\bbS-\bbD\Big)
\end{equation}
and use it to write the $K$th iterate of the Jacobi recursion as
\begin{equation}\label{eqn_jacobi_filter}
   \bbu_{pK}  =  \beta_p \sum_{k=0}^{K-1} \bbR^k(\gamma_p) \bbx \,+\, \bbR^K (\gamma_p) \bbx .
\end{equation}
For a convergent Jacobi recursion, signal $\bbu_{pK}$ in \eqref{eqn_jacobi_filter} converges to the output $\bbu_p$ of the single-pole filter in \eqref{eqn_FIR_direct}. Truncating \eqref{eqn_jacobi_filter} at a finite $K$ yields an approximation in which single-pole filters are written as polynomials on the shift operator $\bbR(\gamma_p)$. I.e., a single-pole filter is approximated as a convolutional filter of order $K$ [cf. \eqref{eqn_gcnn}] in which the shift operator of the graph $\bbS$ is replaced by the shift operator $\bbR(\gamma_p)$ defined in \eqref{eqn_jacobi_shift}. This convolutional filter uses parameters $\beta_p$ for $k=0,\ldots,K-1$ and $1$ for $k=K$.

\medskip\noindent{\bf Jacobi ARMA filters and Jacobi ARMA GNNs.} Assuming we use Jacobi iterations to approximate all single-pole filters in \eqref{eqn_arma_partial_out} and that we truncate all of these iterations at $K$, we can write ARMA filters as
\begin{equation}\label{eqn_arma_jacobi_filter_1}
   \bbA(\bbS) =  \sum_{p = 1}^P \bbH_K\big( \bbR (\gamma_p)\big)
                          + \sum_{k = 0}^{K} \alpha_k\bbS^k .
\end{equation}
where $\bbH_K(\bbR (\gamma_p))$ is a $K$ order Jacobi approximation of the ARMA filter, which, as per \eqref{eqn_jacobi_filter} is given by
\begin{equation}\label{eqn_arma_jacobi_filter_2}
   \bbH_K\big( \bbR (\gamma_p)\big)
      =  \beta_p \sum_{k=0}^{K-1} \bbR^k(\gamma_p) \,+\, \bbR^K (\gamma_p) .
\end{equation}
A Jacobi ARMA filter of order $(P,K)$ is defined by \eqref{eqn_arma_jacobi_filter_1} and \eqref{eqn_arma_jacobi_filter_2}. The order $P$ represents the number of poles in the filter and the order $K$ the number of Jacobi iterations we consider appropriate to properly approximate individual single-pole filters. Notice the number of taps $K$ in the filter $\sum_{k = 0}^{K} \alpha_k\bbS^k$ need not be the same as the number of Jacobi iterations used in \eqref{eqn_arma_jacobi_filter_2}. But we use the same to avoid complicating notation.

For sufficiently large $K$ \eqref{eqn_arma_jacobi_filter_1}-\eqref{eqn_arma_jacobi_filter_2}, \eqref{eqn_arma_partial_out}, and \eqref{eqn_arma} are all equivalent expressions of ARMA filters of orders $(P,Q)$. We could train parameters using either of these equivalent expressions but we advocate for the use \eqref{eqn_arma_jacobi_filter_1}-\eqref{eqn_arma_jacobi_filter_2} as no inversions are necessary except for the elementary inversion of the diagonal matrix $(\bbD - \gamma_p\bbI)$. It is interesting to note that in this latter form ARMA filters are reminiscent of the convolutional filters in \eqref{eqn_gcnn} but the similarity is superficial. In \eqref{eqn_gcnn}, we train $K+1$ parameters $a_k$ that multiply shift operator powers $\bbS^k$. In \eqref{eqn_arma_jacobi_filter_1}-\eqref{eqn_arma_jacobi_filter_2} we also train $K+1$ parameters of this form in the filter $\sum_{k = 0}^{K} \alpha_k\bbS^k$ but this is in addition to the parameters $\beta_p$ and $\gamma_p$ of each of the single-pole filter approximations $\bbH_K( \bbR (\gamma_p))$. These single-pole filters are themselves reminiscent of the convolutional filters in \eqref{eqn_gcnn} but the similarity is again superficial. Instead of parameters $a_k$ that multiply shift operator powers $\bbS^k$, the filters in \eqref{eqn_arma_jacobi_filter_2} train a parameters $\gamma_p$ which represents a constant that is subtracted from the diagonal entries of the shift operators $\bbS$. The fact this is equivalent to an ARMA filter suggests \eqref{eqn_arma_jacobi_filter_1}-\eqref{eqn_arma_jacobi_filter_2} may help designing more discriminative filters. We corroborate in Section \ref{sec:nr} that GNNs using \eqref{eqn_arma_jacobi_filter_1}-\eqref{eqn_arma_jacobi_filter_2} outperform GNNs that utilize filters \eqref{eqn_gcnn}.

An ARMA GNN has $(2P + K +1)F^2$ parameters per layer and a computational complexity of order $\ccalO\big(F^2P(MK+N)	\big)$. This decomposes as $\ccalO(PN)$ to invert the diagonal matrices $(\bbD - \gamma_p\bbI_N)$; $\ccalO(PM)$ to scale the nonzeros of $(\bbS - \bbD)$ by the inverse diagonal; $\ccalO(PKM)$ to obtain the outputs of Jacobi ARMA filters \eqref{eqn_arma_jacobi_filter_1} of order $K$.

\medskip\noindent{\bf ARMA GNNs as EdgeNets.} ARMA GNNs (ARMANets) are another subclass of the EdgeNet. To see this, consider that each shift operator $\bbR(\gamma_p)$ in \eqref{eqn_jacobi_shift} shares the support with $\bbI_N + \bbS$. Hence, we can express the graph filter in \eqref{eqn_arma_jacobi_filter_1} as the union of $P+1$ edge varying graph filters. The first $P$ of these filters have parameters matrices of the form
\[
   \bbPhi^{(k:0)}_p =\left\{
                \begin{array}{ll}\label{}
                  \beta_p\bbR^k(\gamma_p),&k = 0, \ldots, K-1\\
                  \bbR^K(\gamma_p), &k = K
                \end{array}
              \right.
  \]
while the last filter captures the direct-term with edge varying parameter matrices $\bbPhi^{(k:0)} = \alpha_k\bbS^K$ [cf. Section~\ref{sec:gcnn}]. The union of these edge varying filter has the expression
\begin{equation}
\bbA(\bbS) = \sum_{k = 0}^K\bigg(\sum_{p = 1}^P\bbPhi_p^{(k:0)} + \bbPhi^{(k:0)}		\bigg)\bbx
\end{equation}
which by grouping further the terms of the same order $k$ leads to a single edge varying graph filter of the form in \eqref{eqn_evFG}.

%
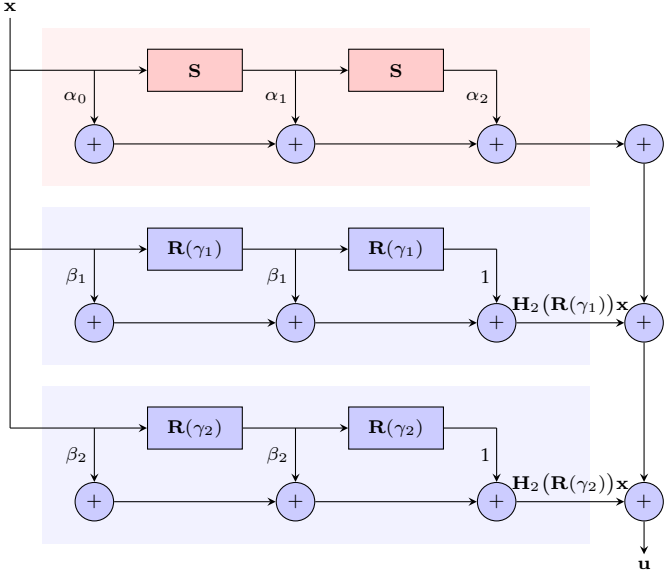
\begin{figure}[!t]
\centering

\def \thisplotscale {1.4}
\def \unit {\thisplotscale cm}

\tikzstyle {Phi base} = [rectangle, 
                         thin,
                         minimum width =  0.9*\unit, 
                         minimum height = 0.4*\unit, 
                         anchor = west,
                         draw]

\tikzstyle {sum} = [circle, 
                    thin,
                    minimum width  = 0.3*\unit, 
                    minimum height = 0.3*\unit, 
                    anchor = center,
                    draw,
                    fill = blue!20]

\def \deltax {1.0}
\def \deltay {0.7}
\def \deltafilter {0.3}
\def \sumshift {0.5}

{\fontsize{7}{7}\selectfont\begin{tikzpicture}[x = 1*\unit, y = 1*\unit]
    
\pgfdeclarelayer{bg}    
\pgfsetlayers{bg,main}  


   \node (input) [above] {$\bbx$};    
    

   \tikzstyle {Phi} = [Phi base, fill = red!20];

   \path (input) ++ (0.3, -0.6) node (0) [] {};    
    
   \path (0)      ++ (1.00*\deltax, 0) node (1) [Phi] {$\bbS$};
   \path (1.east) ++ (1.00*\deltax, 0) node (2) [Phi] {$\bbS$};

   \path (1.west) ++ (-0.5*\deltax, -\deltay) node (sum0) [sum] {$+$};
   \path (1.east) ++ ( 0.5*\deltax, -\deltay) node (sum1) [sum] {$+$};
   \path (2.east) ++ ( 0.5*\deltax, -\deltay) node (sum2 filter1) [sum] {$+$};

   \path[-stealth] (input |- 0) edge [above,       pos=0.45] node {} (1);	
   \path[-stealth] (1)          edge [above right, pos=0.00] node {} (2);	
   \path[-]        (2)          edge [above right, pos=0.00] node {} (2 -| sum2 filter1);	

   \path[-stealth, draw] (sum0 |- 0)         -- node [left, pos=0.5 ] {$\alpha_0$} (sum0);	
   \path[-stealth, draw] (sum1 |- 1)         -- node [left, pos=0.5 ] {$\alpha_1$} (sum1);	
   \path[-stealth, draw] (sum2 filter1 |- 2) -- node [left, pos=0.5 ] {$\alpha_2$} (sum2 filter1);	

   \path[-stealth, draw] (sum0) -- (sum1);	
   \path[-stealth, draw] (sum1) -- (sum2 filter1);	
   
   \begin{pgfonlayer}{bg}
      \path [] (0 |- 1.north) ++ ( 0.0, 0.2) node 
               [fill = red!5, 
                anchor = north west,
                minimum width  = 5.2*\unit, 
                minimum height = 1.5*\unit] {} ; 
   \end{pgfonlayer}


   \tikzstyle {Phi} = [Phi base, fill = blue!20];

   \path (0) ++  (0, -\deltay) ++  (0, -\deltay) ++  (0, -\deltafilter) node (0) [] {};    
    
   \path (0)      ++ (1.00*\deltax, 0) node (1) [Phi] {$\bbR (\gamma_1)$};
   \path (1.east) ++ (1.00*\deltax, 0) node (2) [Phi] {$\bbR (\gamma_1)$};

   \path (1.west) ++ (-0.5*\deltax, -\deltay) node (sum0) [sum] {$+$};
   \path (1.east) ++ ( 0.5*\deltax, -\deltay) node (sum1) [sum] {$+$};
   \path (2.east) ++ ( 0.5*\deltax, -\deltay) node (sum2 filter2) [sum] {$+$};

   \path[-stealth] (input |- 0) edge [above,       pos=0.45] node {} (1);	
   \path[-stealth] (1)          edge [above right, pos=0.00] node {} (2);	
   \path[-]        (2)          edge [above right, pos=0.00] node {} (2 -| sum2 filter2);	

   \path[-stealth, draw] (sum0 |- 0) -- node [left, pos=0.5 ] {$\beta_1$} (sum0);	
   \path[-stealth, draw] (sum1 |- 1) -- node [left, pos=0.5 ] {$\beta_1$} (sum1);	
   \path[-stealth, draw] (sum2 filter2 |- 2) -- node [left, pos=0.5 ] {1} (sum2 filter2);	

   \path[-stealth, draw] (sum0) -- (sum1);	
   \path[-stealth, draw] (sum1) -- (sum2 filter2);	
   
   \begin{pgfonlayer}{bg}
      \path [] (0 |- 1.north) ++ ( 0.0, 0.2) node 
               [fill = blue!5, 
                anchor = north west,
                minimum width  = 5.2*\unit, 
                minimum height = 1.5*\unit] {} ; 
   \end{pgfonlayer}


   \path (0) ++  (0, -\deltay) ++  (0, -\deltay) ++  (0, -\deltafilter) node (0) [] {};    
    
   \path (0)      ++ (1.00*\deltax, 0) node (1) [Phi] {$\bbR (\gamma_2)$};
   \path (1.east) ++ (1.00*\deltax, 0) node (2) [Phi] {$\bbR (\gamma_2)$};

   \path (1.west) ++ (-0.5*\deltax, -\deltay) node (sum0) [sum] {$+$};
   \path (1.east) ++ ( 0.5*\deltax, -\deltay) node (sum1) [sum] {$+$};
   \path (2.east) ++ ( 0.5*\deltax, -\deltay) node (sum2 filter3) [sum] {$+$};

   \path[-stealth] (input |- 0) edge [above,       pos=0.45] node {} (1);	
   \path[-stealth] (1)          edge [above right, pos=0.00] node {} (2);	
   \path[-]        (2)          edge [above right, pos=0.00] node {} (2 -| sum2 filter3);	

   \path[-stealth, draw] (sum0 |- 0) -- node [left, pos=0.5 ] {$\beta_2$} (sum0);	
   \path[-stealth, draw] (sum1 |- 1) -- node [left, pos=0.5 ] {$\beta_2$} (sum1);	
   \path[-stealth, draw] (sum2 filter3 |- 2) -- node [left, pos=0.5 ] {1} (sum2 filter3);	
 
   \path[-stealth, draw] (sum0) -- (sum1);	
   \path[-stealth, draw] (sum1) -- (sum2 filter3);	

   \begin{pgfonlayer}{bg}
      \path [] (0 |- 1.north) ++ ( 0.0, 0.2) node 
               [fill = blue!5, 
                anchor = north west,
                minimum width  = 5.2*\unit, 
                minimum height = 1.5*\unit] {} ; 
   \end{pgfonlayer}


   \path[-] (sum2 filter1) ++ (1.4, 0)  node (out sum 1) [sum] {$+$};	
   \path[-] (sum2 filter2 -| out sum 1) node (out sum 2) [sum] {$+$};	
   \path[-] (sum2 filter3 -| out sum 1) node (out sum 3) [sum] {$+$};
   
   \path[-stealth] (sum2 filter1) edge [above] 
                   node {} (out sum 1);	
   \path[-stealth] (sum2 filter2) edge [above] 
                   node {$\bbH_2\big( \bbR (\gamma_1)\big)  \bbx$} (out sum 2);	
   \path[-stealth] (sum2 filter3) edge [above] 
                   node {$\bbH_2\big( \bbR (\gamma_2)\big)  \bbx$} (out sum 3);	

   \path[draw, -stealth] (out sum 1) -- (out sum 2);	
   \path[draw, -stealth] (out sum 2) -- (out sum 3);
   \path[draw, -stealth] (out sum 3) -- ++ (0, -0.5) node [below] {$\bbu$};


   \path[draw, -] (input) -- (input |- 0);


\end{tikzpicture}}
\caption{Jacobi Autoregressive Moving Average Filter. The input signal $\bbx$ is processed by a parallel bank of filters. One of this filters is a convolutional filter of the form in \eqref{eqn_gcnn} operating w.r.t. the shift operator $\bbS$ (highlighted in red). The remaining filters operate w.r.t. scaled shift operators [cf. \eqref{eqn_jacobi_shift}] (highlighted in blue). All filter outputs are summed together to yield the overall Jacobi ARMA output.}
\label{fig.jacARMA}
\end{figure}

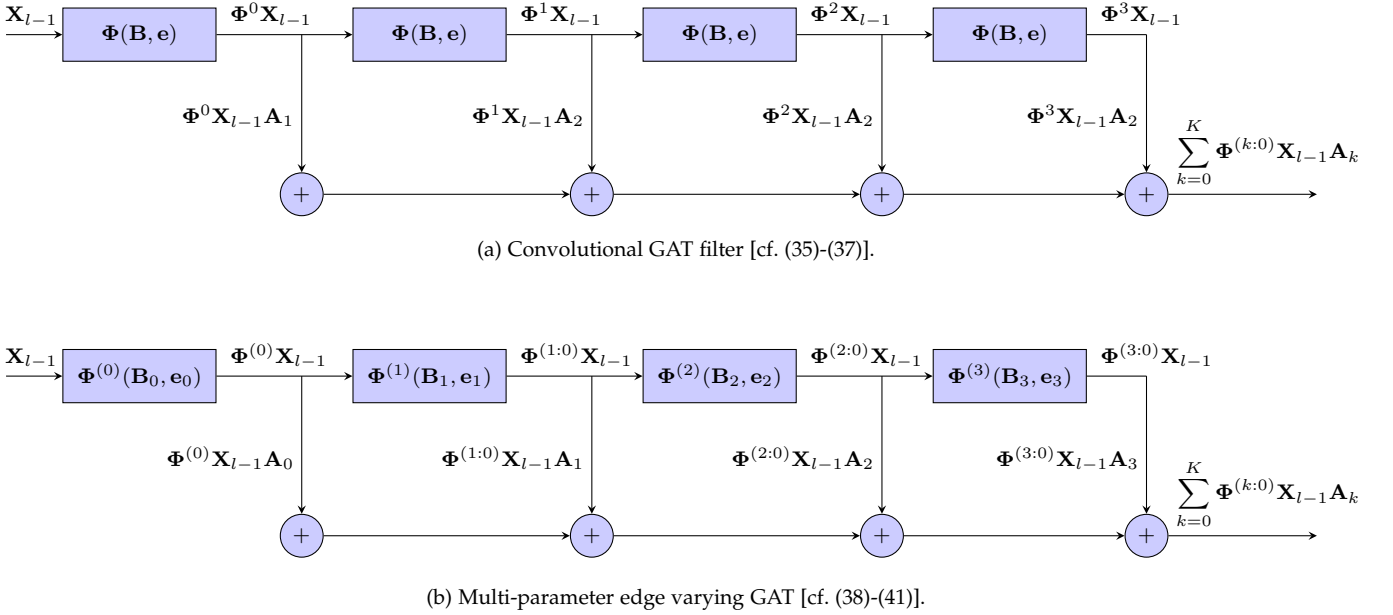
\begin{figure*}[!t]%
\centering

\def \thisplotscale {1.77}
\def \unit {\thisplotscale cm}

\tikzstyle {Phi} = [rectangle, 
                    thin,
                    minimum width = 1.15*\unit, 
                    minimum height = \sumshift*\unit, 
                    anchor = west,
                    draw,
                    fill = blue!20]

\tikzstyle {sum} = [circle, 
                    thin,
                    minimum width  = 0.3*\unit, 
                    minimum height = 0.3*\unit, 
                    anchor = center,
                    draw,
                    fill = blue!20]

\def \deltax {1.6}
\def \deltay {1.2}
\def \deltagat {1.2}
\def \sumshift {0.4}

{\footnotesize\begin{tikzpicture}[x = 1*\unit, y = 1*\unit]
    
\node (first) [] {};    
    
\path (first) ++ (0.30*\deltax, 0) node (0) [Phi] {$\bbPhi(\bbB, \bbe)$};
\path (0)     ++ (1.00*\deltax, 0) node (1) [Phi] {$\bbPhi(\bbB, \bbe)$};
\path (1)     ++ (1.00*\deltax, 0) node (2) [Phi] {$\bbPhi(\bbB, \bbe)$};
\path (2)     ++ (1.00*\deltax, 0) node (3) [Phi] {$\bbPhi(\bbB, \bbe)$};

\path (3.east) ++ (0.7*\sumshift*\deltax, 0) node [anchor=west] (last) [] {};


\path (0.east) ++ (\sumshift*\deltax, -\deltay) node (sum0) [sum] {$+$};
\path (1.east) ++ (\sumshift*\deltax, -\deltay) node (sum1) [sum] {$+$};
\path (2.east) ++ (\sumshift*\deltax, -\deltay) node (sum2) [sum] {$+$};
\path (3.east) ++ (0.7*\sumshift*\deltax, -\deltay) node (sum3) [sum] {$+$};

\path[-stealth] (first) edge [above, pos=0.45] node {$\bbX_{l-1}$}                 (0);	
\path[-stealth] (0)     edge [above right, pos=0.0] node {$\ \bbPhi^{0}\bbX_{l-1}$}   (1);	
\path[-stealth] (1)     edge [above right, pos=0.0] node {$\ \bbPhi^{1}\bbX_{l-1}$} (2);	
\path[-stealth] (2)     edge [above right, pos=0.0] node {$\ \bbPhi^{2}\bbX_{l-1}$} (3);
\path[-]        (3)     edge [above right, pos=0.0] node {$\ \bbPhi^{3}\bbX_{l-1}$} (last);				


\path[-stealth, draw] (sum0 |- 0) -- node [left, pos=0.6 ] {$\ \bbPhi^{0}  \bbX_{l-1}\bbA_1$} (sum0);	
\path[-stealth, draw] (sum1 |- 1) -- node [left, pos=0.6 ] {$\ \bbPhi^{1}\bbX_{l-1}\bbA_2$} (sum1);	
\path[-stealth, draw] (sum2 |- 2) -- node [left, pos=0.6 ] {$\ \bbPhi^{2}\bbX_{l-1}\bbA_2$} (sum2);	
\path[-stealth, draw] (sum3 |- 3) -- node [left, pos=0.6 ] {$\ \bbPhi^{3}\bbX_{l-1}\bbA_2$} (sum3);	

\path[-stealth, draw] (sum0) -- (sum1);	
\path[-stealth, draw] (sum1) -- (sum2);	
\path[-stealth, draw] (sum2) -- (sum3);	

\path[-stealth] (sum3) edge [above right, pos=0.0] node 
                {$\displaystyle{\sum_{k=0}^K  \bbPhi^{(k:0)} \bbX_{l-1} \bbA_{k}}$} ++ (0.8*\deltax, 0);

\end{tikzpicture}}\\ \medskip
{\footnotesize (a) Convolutional GAT filter [cf. \eqref{eqn_polyGat}-\eqref{eqn_nonlinCoeff}]. \\}
\vspace{10mm}

\def \thisplotscale {1.77}
\def \unit {\thisplotscale cm}

\tikzstyle {Phi} = [rectangle, 
                    thin,
                    minimum width = 1.15*\unit, 
                    minimum height = \sumshift*\unit, 
                    anchor = west,
                    draw,
                    fill = blue!20]

\tikzstyle {sum} = [circle, 
                    thin,
                    minimum width  = 0.3*\unit, 
                    minimum height = 0.3*\unit, 
                    anchor = center,
                    draw,
                    fill = blue!20]

\def \deltax {1.6}
\def \deltay {1.2}
\def \deltagat {1.2}
\def \sumshift {0.4}

{\footnotesize\begin{tikzpicture}[x = 1*\unit, y = 1*\unit]
    
\node (first) [] {};    
    
\path (first) ++ (0.30*\deltax, 0) node (0) [Phi] {$\bbPhi^{(0)}(\bbB_0, \bbe_0)$};
\path (0)     ++ (1.00*\deltax, 0) node (1) [Phi] {$\bbPhi^{(1)}(\bbB_1, \bbe_1)$};
\path (1)     ++ (1.00*\deltax, 0) node (2) [Phi] {$\bbPhi^{(2)}(\bbB_2, \bbe_2)$};
\path (2)     ++ (1.00*\deltax, 0) node (3) [Phi] {$\bbPhi^{(3)}(\bbB_3, \bbe_3)$};

\path (3.east) ++ (0.7*\sumshift*\deltax, 0) node [anchor=west] (last) [] {};


\path (0.east) ++ (\sumshift*\deltax, -\deltay) node (sum0) [sum] {$+$};
\path (1.east) ++ (\sumshift*\deltax, -\deltay) node (sum1) [sum] {$+$};
\path (2.east) ++ (\sumshift*\deltax, -\deltay) node (sum2) [sum] {$+$};
\path (3.east) ++ (0.7*\sumshift*\deltax, -\deltay) node (sum3) [sum] {$+$};

\path[-stealth] (first) edge [above, pos=0.45] node {$\bbX_{l-1}$}                 (0);	
\path[-stealth] (0)     edge [above right, pos=0.0] node {$\ \bbPhi^{(0)}\bbX_{l-1}$}   (1);	
\path[-stealth] (1)     edge [above right, pos=0.0] node {$\ \bbPhi^{(1:0)}\bbX_{l-1}$} (2);	
\path[-stealth] (2)     edge [above right, pos=0.0] node {$\ \bbPhi^{(2:0)}\bbX_{l-1}$} (3);
\path[-]        (3)     edge [above right, pos=0.0] node {$\ \bbPhi^{(3:0)}\bbX_{l-1}$} (last);				


\path[-stealth, draw] (sum0 |- 0) -- node [left, pos=0.6 ] {$\ \bbPhi^{(0)}  \bbX_{l-1}\bbA_0$} (sum0);	
\path[-stealth, draw] (sum1 |- 1) -- node [left, pos=0.6 ] {$\ \bbPhi^{(1:0)}\bbX_{l-1}\bbA_1$} (sum1);	
\path[-stealth, draw] (sum2 |- 2) -- node [left, pos=0.6 ] {$\ \bbPhi^{(2:0)}\bbX_{l-1}\bbA_2$} (sum2);	
\path[-stealth, draw] (sum3 |- 3) -- node [left, pos=0.6 ] {$\ \bbPhi^{(3:0)}\bbX_{l-1}\bbA_3$} (sum3);	

\path[-stealth, draw] (sum0) -- (sum1);	
\path[-stealth, draw] (sum1) -- (sum2);	
\path[-stealth, draw] (sum2) -- (sum3);	

\path[-stealth] (sum3) edge [above right, pos=0.0] node 
                {$\displaystyle{\sum_{k=0}^K  \bbPhi^{(k:0)} \bbX_{l-1} \bbA_{k}}$} ++ (0.8*\deltax, 0);

\end{tikzpicture}}\\ \medskip
{\footnotesize (b) Multi-parameter edge varying GAT [cf. \eqref{eqn_gat_edge_varying}-\eqref{eqn_nonlinCoeff_edge_varying}].}
\caption{Higher-order Graph Attention Filters. (a) Graph convolutional attention filter. The input features $\bbX_{l-1}$ are shifted by the same edge varying shift operator $\bbPhi(\bbB, \bbe)$ and weighted by different parameter matrices $\bbA_k$. The edge varying parameters in all $\bbPhi(\bbB, \bbe)$ are parameterized by the same matrix $\bbB$ and vector $\bbe$ following the attention mechanism. (b) Edge varying GAT filter. The input features $\bbX_{l-1}$ are shifted by different edge varying shift operators $\bbPhi^{(k)}(\bbB_k, \bbe_k)$ and weighted by different parameter matrices $\bbA_k$. The edge varying parameters in the different $\bbPhi^{(k)}(\bbB_k, \bbe_k)$ are parameterized by a different matrix $\bbB_k$ and vector $\bbe_k$ following the attention mechanism.}%
\label{fig:GATmain}%
\end{figure*}


ARMANet provides an alternative parameterization of the EdgeNet that is different from that of the other polynomial convolutional filters in \eqref{eqn_gcnn}. In particular, ARMANets promote to use multiple polynomial filters of smaller order (i.e., the number of Jacobi iterations) with shared parameters between them. Each of the filters $\bbH_K(\bbR(\gamma_p))$ depends on two parameters $\beta_p$ and $\gamma_p$. We believe this parameter sharing among the different orders and the different nodes is the success behind the improved performance of the ARMA GNN compared with the single polynomial filters in \eqref{eqn_gcnn}. Given also the hybrid solutions developed in Section~\eqref{subsec:nodeDep} for the polynomial filters, a direction that may attain further improvements is that of ARMA GNN architectures with controlled edge variability.

{ARMANet generalizes the architecture in \cite{levie2017cayleynets} where instead of restricting the polynomials in \eqref{eqn_arma} to Cayley polynomials, it allows the use of general polynomials. Iterative approaches to implement an ARMA graph filter \cite{isufi2017autoregressive,isufi2017autoregressivest} have been recently used to develop GNNs that resemble this recursion \cite{bianchi2019graph, wijesinghe2019dfnets}. However, both \cite{bianchi2019graph, wijesinghe2019dfnets} do not implement an ARMA layer like (23) does. Instead, their propagation rule can be either seen as that used in graph recurrent neural networks \cite[eq. (5)]{ruiz2020gated} but with a constant input or conceptually similar to that used in iterative sparse coding \cite{gregor2010learning}.}

\section{\!\!\!\!Graph Convolutional Attention Networks}
\label{sec:gat}

%
{Graph convolutional neural networks parameterize the EdgeNet by using a fixed underlying shift operator $\mathbf{S}$ and learning only the filter parameters. However, this shift operator may often be estimated from data disjointly from the GNN task, or its specific weights may be unknown. In these instances, we can use the graph attention mechanism \cite{velickovic2017graph} to parameterize the EdgeNet in a way that we learn both the shift operator weights and the convolutional filter parameters for the task at hand. We propose the graph convolutional attention network (GCAT), which utilizes the filters as in \eqref{eqn_gcnn_matrix_notation} but they are convolutional in a layer-specific matrix $\bbPhi_l\!\!=\!\!\bbPhi$ that may be different from the shift operator \!$\bbS$}
%
\begin{equation}\label{eqn_polyGat}
   \bbX_l = \sigma \Bigg( \sum_{k=0}^K  \bbPhi^{k} \bbX_{l-1} \bbA_{k}\Bigg) .
\end{equation}
Note $\bbA_{k}= \bbA_{lk}$ and $\bbPhi=\bbPhi_l$ are layer-dependent but we omit the layer index to simplify notation. Since matrix $\bbPhi$ shares the sparsity pattern of $\bbS$, \eqref{eqn_polyGat} defines a GNN as per \eqref{eqn_gnn_multiple_feature}. Matrix $\bbPhi$ is learned from the features $\bbX_{l-1}$ passed from layer $l-1$ following the attention mechanism \cite{velickovic2017graph}. {Specifically, we consider a trainable matrix $\bbB \in\reals^{F_{l-1}\times F_l}$ and vector $\bbe\in\reals^{2F_l}$, and compute the edge scores}
\begin{equation}\label{eqn_gat_edge_scores}
   \alpha_{ij}
     = \sigma\bigg( \bbe^\top \Big[\big[\bbX_{l-1} \bbB\big]_i,\,
                                \big[\bbX_{l-1} \bbB\big]_j\Big]^\top \bigg)
\end{equation}
for all edges $(i,j) \in \ccalE$. In \eqref{eqn_gat_edge_scores}, we start with the vector of features $\bbX_{l-1}$ and mix them as per the parameters in $\bbB$. This produces a collection of graph signals $\bbX_{l-1} \bbB$ in which each node $i$ has $F_l$ features that correspond to the $i$th row $[\bbX_{l-1} \bbB]_i$ of the product matrix $\bbX_{l-1} \bbB$. The features at node $i$ are concatenated with the features of node $j$ and the resulting vector of $2F_l$ components is multiplied by vector $\bbe$. This product produces the score $\alpha_{ij}$ after passing through the nonlinearity $\sigma(\cdot)$. {Note that $\bbB=\bbB_l$, $\bbe=\bbe_l$ are global parameters for all scores $\alpha_{ij}=\alpha_{lij}$ and depend on the layer index $l$.} As is the case of $\bbA_{k}$ and $\bbPhi$  in \eqref{eqn_polyGat}, we omitted this index for simplicity.

The score $\alpha_{ij}$ could be used directly as an entry for the matrix $\bbPhi$ but to encourage attention sparsity we pass $\alpha_{ij}$ through a local soft maximum operator
\begin{equation}\label{eqn_nonlinCoeff}
   \Phi_{ij} 
             =  \text{exp}\big(\alpha_{ij}\big) \ \ \Bigg(
                \sum_{j'\in\ccalN_i \cup i}\text{exp}\big(\alpha_{ij'} \big) \Bigg)^{-1}.
\end{equation}
The soft maximum assigns edge weights $\Phi_{ij}$ close to 1 to the largest of the edge scores $\alpha_{ij}$ and weights $\Phi_{ij}$ close to 0 to the rest. See also Figure~\ref{fig:GATmain} (a).

In Section \ref{sec_ev_graph_filters}, we introduced arbitrary edge varying graph filters [cf. \eqref{eqn_evFG}] which we leveraged in Section \ref{sec_gnn} to build edge varying GNNs [cf. \eqref{eqn_gnn_single_feature} - \eqref{eqn_gnn_multiple_feature}]. In Section \ref{sec:gcnn}, we pointed out that edge varying graph filters left too many degrees of freedom in the learning parametrization; a problem that we could overcome with the use of graph convolutional filters [cf. \eqref{eqn_gcnn}]. The latter suffer from the opposite problem as they may excessively constrict the GNN. GATs provide a solution of intermediate complexity. Indeed, the filters in \eqref{eqn_polyGat} allow us to build a GNN with convolutional graph filters where the shift operator $\bbPhi$ is learned ad hoc in each layer to represent the required abstraction between nodes. The edges of this shift operator try to choose neighbors whose values should most influence the computations at a particular node. This is as in any arbitrary edge varying graph filter but the novelty of GATs is to reduce the number of learnable parameters by tying edge values to matrix $\bbB$ and vector $\bbe$ ---observe that in \eqref{eqn_gat_edge_scores} $\bbe$ is the same for all edges. Thus, the computation of scores $\alpha_{ij}$ depends on the $F_{l-1}\times F_l$ parameters in $\bbB$ and the $2F_l$ parameters in $\bbe$. This is of order no more than $F^2$ if we make $F=\max_{l} F_l$. It follows that for the GAT in \eqref{eqn_polyGat} the number of learnable parameters is at most $F^2 + 2F + F^2(K + 1)$, which depends on design choices and is independent of the number of edges. We point out that since $\bbPhi$ respects the graph sparsity, the computational complexity of implementing \eqref{eqn_polyGat} and its parameterization is of order $\ccalO(F(NF+ KM))$. \vspace{-3mm}


\begin{table*}[t]
\centering
\caption{Properties of Different Graph Neural Network Architectures. The parameters and complexity are considered per layer. Architectures in bold are proposed in this work. Legend: $N$- number nodes; $M$- number of edges; $F$- maximum number of features; $K$- recursion order; $b$ - dimension of the Kernel in \eqref{eq.EV_respfinProp}; $B$ - number of blocks in \eqref{eq.nvGCNN}; $\ccalI$ - the set of important nodes in \eqref{eq.nvGCNN} and \eqref{eq.nvGCNN1}; $M_{\ccalI}$ - total neighbors for the nodes in $\ccalI$; $P$ - parallel J-ARMANet branches; $R$ - parallel attention branches; $^*$Self-loops are not considered. $^{**}$The eigendecomposition cost $\ccalO(N^3)$ is computed once.}
\begin{tabular}{c c c c}
\Xhline{4\arrayrulewidth}		
  Architecture & Expression & Order parameters$^*$ $\ccalO(\cdot)$ & Order of complexity$^{*, **}$ $\ccalO(\cdot)$\\
  \hline
  \rowcolor{Gray}
Fully connected & n/a & $N^2F^2$  & $N^2F^2$\\
\textbf{Edge varying} & Eq.~\eqref{eqn_gnn_multiple_feature} & $K(M+N)F^2$  & $KF^{2}(M+N)$\\
\rowcolor{Gray}
GCNN \cite{defferrard2016convolutional, kipf2016semi,du2017topology, gama2018convolutional} & Eq. \eqref{eqn_gcnn_matrix_notation} & $KF^2$ & $KF^2M$\\
Node varying \cite{gama2018convolutionalNV} & Eq. \eqref{eq.nvGCNN} &$BKF^2$ & $KF^2M$\\
\rowcolor{Gray}
\textbf{Hybrid edge varying}& Eq. \eqref{eq.nvGCNN1} & $(|\ccalI| +  KM_\ccalI)F^2$ & $KF^{2}(M+N)$\\
\textbf{Spec. edge varying GNN}$^{**}$ & Eq.~\eqref{eq.EV_respfinProp} & $KbF^2$ & $KF^2(M+N)$\\
\rowcolor{Gray}
Spec. Kernel GCNN$^{**}$ \cite{bruna2013spectral} & Eq.~\eqref{eq.EV_respfinProp} for $K = 1$ & $bF^2$ & $N^2F^2$\\
\textbf{ARMANet} & Eq. \eqref{eqn_arma}-\eqref{eqn_arma_jacobi_filter_2} & $(P + K)F^2$ & $F^2P(MK+N)$\\
\rowcolor{Gray}
\textbf{GCAT} & Eq. \eqref{eqn_polyGat} & $RF^2K$ & $R(NF^2+ KFM)$\\
GAT \cite{velickovic2017graph} & Eq. \eqref{eqn_polyGat} for $K = 1$ & $RF^2 $ & $R(NF^2+FM)$\\
\rowcolor{Gray}
\textbf{Edge varying GAT} & Eq. \eqref{eqn_gat_edge_varying} & $RKF^2$  &$RK(NF^2 + MF)$\\
\Xhline{4\arrayrulewidth}
\end{tabular}
\label{tab:sum}\vskip-4mm
\end{table*}

%
\subsection{Edge varying GAT networks}

The idea of using attention mechanisms to estimate entries of a shift operator $\bbPhi$ can be extended to estimate entries $\bbPhi^{(k:0)}$ of an edge varying graph filter. To be specific, we propose to implement a generic GNN as defined by recursion \eqref{eqn_gnn_multiple_feature} which we repeat here for ease of reference
\begin{equation}\label{eqn_gat_edge_varying}
   \bbX_l = \sigma \Bigg( \sum_{k=0}^K  \bbPhi^{(k:0)} \bbX_{l-1} \bbA_{k}\Bigg) .
\end{equation}
Further recall each edge varying filter parameter matrix $\bbPhi^{(k:0)}$ is itself defined recursively as [cf. \eqref{eqn_evFG}]
\begin{equation}\label{eqn_gat_edge_varying_coefficient_recursion}
   \bbPhi^{(k:0)} \ = \  \bbPhi^{(k)}\,\,\bbPhi^{(k-1:0)}
                  \ = \  \prod_{k'=0}^{k} \bbPhi^{(k')} .
\end{equation}
We propose to generalize \eqref{eqn_gat_edge_scores} so that we compute a different matrix $\bbPhi^{(k)}$ for each filter order $k$. Consider then matrices $\bbB_k$ and vectors $\bbe_k$ to compute the edge scores
\begin{equation}\label{eqn_gat_edge_scores_time_varying}
   \alpha_{ij}^{(k)}
     = \sigma\bigg( \bbe_k^T \Big[\big[\bbX_{l-1} \bbB_k\big]_i,\,
                                \big[\bbX_{l-1} \bbB_k\big]_j\Big]^\top \bigg)
\end{equation}
for all edges $(i,j) \in \ccalE$. As in \eqref{eqn_gat_edge_scores}, we could use $\alpha_{ij}^{(k)}$ as edge weights in $\bbPhi^{(k)}$, but to promote attention sparsity we send $\alpha_{ij}^{(k)}$ through a soft maximum function to yield edge scores 
\begin{equation}\label{eqn_nonlinCoeff_edge_varying}
   \Phi_{ij}^{(k)} =  \text{exp}\Big(\alpha_{ij}^{(k)}\Big)  \ \ \Bigg(
                           \sum_{j'\in\ccalN_i \cup i}\text{exp}\Big(\alpha_{ij'}^{(k)}\Big)\Bigg)^{-1}.
\end{equation}
Each of the edge varying matrices $\bbPhi^{(k)}$ for $k = 1, \ldots, K$ is parameterized by the tuple of transform parameters $(\bbB_k, \bbe_k)$. Put simply, we are using a different GAT mechanism for each edge varying matrix $\bbPhi^{(k)}$. These learned matrices are then used to build an edge varying filter to process the features $\bbX_l$ passed from the previous layer -- see Figure \ref{fig:GATmain} (b). The edge varying GAT filter employs $K+1$ transform matrices $\bbB_k$ of dimensions $F_{l} \times F_{l-1}$, $K+1$ vectors $\bbe_k$ of dimensions $2F_l$, and $K+1$ matrices $\bbA_k$ of dimensions $F_l \times F_{l-1}$. Hence, the total number of parameters for the edge varying GAT filter is at most $(K+1) (2F^2 + 2F)$. The computational complexity of the edge varying GAT is of order $\ccalO(KF(NF + M))$.



%
\begin{remark}\label{rmk_existing_gats}\normalfont Graph attention networks first appeared in \cite{velickovic2017graph}. In this paper, \eqref{eqn_gat_edge_scores} and \eqref{eqn_nonlinCoeff} are proposed as an attention mechanism for the signals of neighboring nodes and GNN layers are of the form $\bbX_l = \sigma(\bbPhi_l \bbX_{l-1}\bbA_l)$. The latter is a particular case of either \eqref{eqn_polyGat} or \eqref{eqn_gat_edge_varying} in which only the term $k=1$ is not null. Our observation in this section is that this is equivalent to computing a different graph represented by the shift operator $\bbPhi$. This allows for the generalization to filters of arbitrary order $K$ [cf. \eqref{eqn_polyGat}] and to edge varying graph filters of arbitrary order [cf. \eqref{eqn_gat_edge_varying}]. The approaches presented in this paper can likewise be extended with the multi-head attention mechanism proposed in \cite{velickovic2017graph} to improve the network capacity.
 \end{remark}

\begin{figure*}[!t]%
\centering
\includegraphics[width=.9\textwidth,trim={2cm 0cm 3.5cm .08cm},clip]{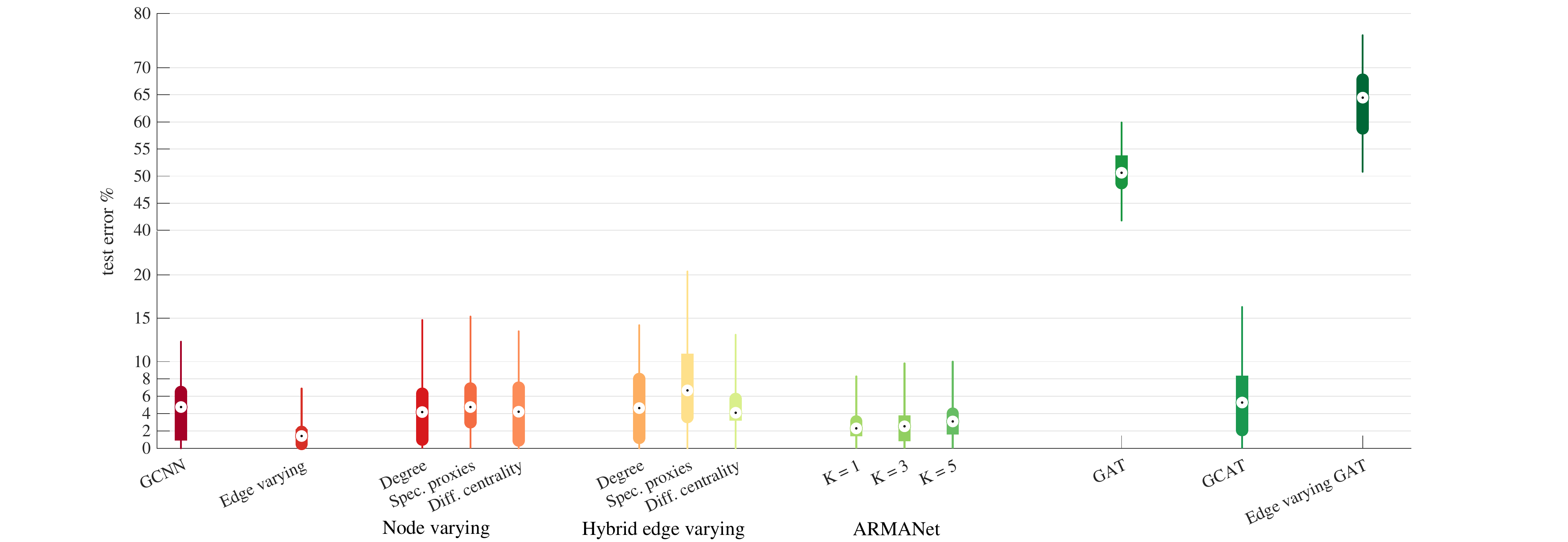}\vskip-2mm
\caption{Source Localization Test Error in the Stochastic Block Model graph. The $y-$axis scale is deformed to improve visibility. The thick bar interval indicates the average performance for different parameter choices (e.g., filter order, attention heads). The circle marker represents the mean value of this interval. The thin line spans an interval of one standard deviation from the average performance. The convolutional-based approaches perform better than attention-based. We attribute the poor performance of the attention techniques to the slow learning rate. Both the GAT and the edge variant GAT required more than $40$ epochs to reach a local minimum. However, the graph convolutional attention network (GCAT) does not suffer from the latter issue leading to faster learning.}%
\label{fig:SourceLoc}\vskip-4mm
\end{figure*}

%
\subsection{Discussions}\label{subsec:disc}

\noindent{\bf Reducing model complexity.} As defined in \eqref{eqn_gat_edge_scores} and \eqref{eqn_gat_edge_scores_time_varying} the attention mechanisms are separate from filtering. To reduce the number of parameters, we can equate the attention matrices $\bbB$ or $\bbB_k$ with the filtering matrices $\bbA_k$. For the GCAT in \eqref{eqn_polyGat}, the original proposal in \cite{velickovic2017graph} is to make $\bbB=\bbA_1$ so that \eqref{eqn_gat_edge_scores} reduces to
\begin{equation}\label{eqn_gat_edge_scores_A}
   \alpha_{ij}
     = \sigma\bigg( \bbe^T \Big[\big[\bbX_{l-1} \bbA_1\big]_i,\,
                                \big[\bbX_{l-1} \bbA_1\big]_j\Big]^\top \bigg).
\end{equation}
For the edge varying GATs in \eqref{eqn_gat_edge_varying}, it is natural to equate $\bbB_k=\bbA_k$ in which case \eqref{eqn_gat_edge_scores_time_varying} reduces to 

\begin{equation}\label{eqn_gat_edge_scores_time_varying_A}
   \alpha_{ij}^{(k)}
     = \sigma\bigg( \bbe_k^T \Big[\big[\bbX_{l-1} \bbA_k\big]_i,\,
                                \big[\bbX_{l-1} \bbA_k\big]_j\Big]^\top \bigg)
\end{equation}
The choice in \eqref{eqn_gat_edge_scores_time_varying_A} removes $(K+1)F^2$ parameters.

\medskip\noindent{\bf Accounting for differences in edge weights in the original shift operator.} The major benefit of the GAT mechanism is to build a GNN without requiring full knowledge of $\bbS$. This is beneficial as it yields a GNN robust to uncertainties in the edge weights. This benefit becomes a drawback when $\bbS$ is well estimated as it renders weights $s_{ij}$ equivalent regardless of their relative values. One possible solution to this latter drawback is to use a weighted soft maximum operator so that the entries of $\bbPhi^{(k)}$ are chosen as
\begin{equation}\label{eqn_nonlinCoeff_GSO}
   \Phi_{ij}^{(k)} =  \text{exp}\Big(s_{ij}\alpha_{ij}^{(k)}\Big)  \ \ \Bigg(
                           \sum_{j'\in\ccalN_i \cup i}
                                \text{exp}\Big( s_{ij'} \alpha_{ij'}^{(k)}\Big)\Bigg)^{-1}.
\end{equation}
Alternatively, we can resort to the use of a hybrid GAT in which we combine a GAT filter of the form in \eqref{eqn_gat_edge_varying} with a regular convolutional filter of the form in \eqref{eqn_gcnn_matrix_notation}
\begin{equation}\label{eq.HybGCAT}
      \bbX_l = \sigma \Bigg( \sum_{k=0}^K  
                   \bbS^{k} \bbX_{l-1} \bbA_{k} + \bbPhi^{(k:0)}\bbX_{l-1} \bbA'_{k} \Bigg) .
\end{equation}
This is the GAT version of the hybrid GNN we proposed in \eqref{eq.nvGCNN1}. The filters in \eqref{eq.HybGCAT} account for both, the GAT learned shifts $\bbPhi^{(k)}$ and the original shift $\bbS$.



\section{Numerical Results}
\label{sec:nr}



This section corroborates the capacity of the different models with numerical results on synthetic and real-world graph signal classification problems. Given the different hyperparameters for the models and the trade-offs (e.g., complexity, number of parameters, radius of local information), we aim to provide insights on which methods exploit better the graph prior for learning purposes rather than achieving the best performance. {In the sequel, we provide a summary of the different architectures, while in the upcoming four section analyze in detail each experiment. Section~\ref{subsec.genobs} provides some general observations. The code used for these simulations can be found at \url{http://github.com/alelab-upenn/graph-neural-networks}.
}

{
\smallskip
\noindent\textbf{Summary of the architectures.} A summary of the different architectures discussed in this paper is provided in Table~\ref{tab:sum}. 
\begin{itemize}
\item Fully connected stands in for the standard multi-layer perceptron neural network that uses no graph prior. 
\item The edge varying stand in for the EdgeNet in the full form in \eqref{eqn_gnn_multiple_feature}, which allocates different parameters per edge and shift and has the most degrees of freedom. 
\item The graph convolutional neural network in \eqref{eqn_gcnn_matrix_notation} has instead shared parameters among all nodes, edges, and shifts.
\item The node varying [cf. \eqref{eq.nvGCNN}] and the hybrid edge varying [cf. \eqref{eq.nvGCNN1}] are the intermediate solutions between the full-form EdgeNet and the GCNN. Both rely on selecting a set of important nodes and allocate different parameters to them. The hybrid edge varying allocates also different parameters to the edges of these important nodes. To the remaining nodes shared parameters as in the GCNN are used.
\item The spectral edge varying GNN is the subclass of the full-form EdgeNet in which filters share the eigenvectors with the shift operator [cf. Prop.~\ref{prop:EVResp}], while the spectral kernel GCNN is the equivalent representation of the GCNN in the spectral domain.
\item ARMANet is the GCNN that we propsoed in Section~\ref{subsec:FIRGCNN}. Contrarily to the polynomial GCNN form [cf. \eqref{eqn_gcnn_matrix_notation}], it has a filter with a rational spectral response rather than polynomial.
\item The graph convolutional attention network and its variants are the architectures we developed in Section~\ref{sec:gat}. The GCAT generalises the popular GAT \cite{velickovic2017graph} to a convolutional filter of higher order such that it can account for multi-hop neighbor information in each layer. The edge varying GAT [cf. \eqref{eqn_gat_edge_varying}] is an alternative to the GCAT that uses edge varying filters together with attention instead of convolutional filters. Lastly, we also discussed two variants of the GCAT and edge varying GAT in Section~\ref{subsec:disc} but several others developed under the EdgeNet framework.
\end{itemize}
From all these architectures, we do not evaluate the fully-connected, the spectral kernel GCNN \cite{bruna2013spectral}, and the spectral edge varying GCNN \eqref{eq.EV_respfinProp} since their computational cost is higher than linear. We also leave to interested readers the extensions discussed in Section~\ref{subsec:disc}. For the remaining solutions, we trained all of them using ADAM with the standard forgetting factors \cite{kingma2014adam}.  \vspace{-4mm}
}


\subsection{Source localization on SBM graphs}

The goal of this experiment is to identify which community in a stochastic block model (SBM) graph is the source of a diffused signal by observing different realizations in different time instants. We considered a connected and undirected graph of $N = 50$ nodes divided into five blocks each representing one community $\{c_1, \ldots, c_5\}$. The intra- and inter-community edge formation probabilities are $0.8$ and $0.2$, respectively. The source node is one of the five nodes ($i_{1}, \ldots, i_{5}$) with the largest degree in the respective community. The source signal $\bbx(0)$ is a Kronecker delta centered at the source node. The source signal is diffused at time $t \in [0, 50]$ as $\bbx(t) = \bbS^t\bbx(0)$, where $\bbS$ is the graph adjacency matrix normalized by the maximum eigenvalue.

The training set is composed of $10240$ tuples of the form $(\bbx(t), c_i)$ for random $t$ and $i \in \{1, \ldots, 5\}$. These tuples are used to train the EdgeNets that are subsequently used to predict the source community $c^\prime_i$ for a testing signal $\bbx^\prime(t)$ again for a random value of $t$. The validation and the test set are both composed of $2560$ tuples ($25\%$ of the training set). The performance of the different algorithms is averaged over ten different graph realizations and ten data splits, for a total of $100$ Monte-Carlo iterations. ADAM is run for $40$ epochs with batches of $100$ samples and learning rate $10^{-3}$.

\medskip\noindent\textbf{Architecture parameters.} For this experiment, we compared $14$ different architectures. All architectures comprise the cascade of a graph filtering layer with ReLU nonlinearity and a fully connected layer with softmax nonlinearity. The architectures are: $i)$ the edge varying GNN \eqref{eqn_gnn_multiple_feature}; $ii)$ the GCNN \eqref{eqn_gcnn_matrix_notation};  $iii)$ three node varying GNNs \eqref{eq.nvGCNN}, where the five important nodes are selected based on $iii-a)$ maximum degree; $iii-b)$ spectral proxies \cite{anis2016efficient}, which ranks the nodes according to their contribution to different frequencies; $iii-c)$ diffusion centrality (see Appendix~\ref{sec:AppdiffCentr}); $iv)$ three node dependent edge varying GNNs, where the five important nodes $\ccalB$ are selected similalr to the node varying case; $v)$ three ARMANets \eqref{eqn_arma_jacobi_filter_1} with Jacobi iterations $v-a)$ $K = 1$; $v-b)$ $K = 3$; $v-c)$ $K = 5$; $vi)$ the GAT network from \cite{velickovic2017graph}; $vii)$ the GCAT network \eqref{eqn_polyGat}; and $viii)$ the edge varying GAT \eqref{eqn_gat_edge_varying} network.


\begin{table*}[!t]
\centering
\caption{Source Localization Test Error in Facebook Subnetwork. The goal is to grid-search the parameters to achieve a mean error of at most $2\%$. For the architectures that did not achieve this criterion, the minimum error is reported.}\vskip-.25cm
\begin{tabular}{l c c | c c c c}
\Xhline{4\arrayrulewidth}
Architecture & mean & std. dev. & order & attention heads & epochs & learning rate\\
~ & ~ & ~  & $\{1,2, 3, 4, 5\}$ & $\{1,2, 3,4, 5\}$ & \{10, 20, 40, 100\}& $\{10^{-2}, 10^{-3}\}$\\
\hline
  \rowcolor{Gray}
GCNN & $4.0\%$ & $13.0\%$ &  $3$ & n/a & $100$ & $10^{-3}$\\
Edge varying & $\mathbf{1.5\%}$ & $8.4\%$ &  $1$ & n/a& $10$ & $10^{-3}$\\
  \rowcolor{Gray}
Node varying & $6.0\%$ & $15.8\%$ &  $3$ & n/a & $20$ & $10^{-3}$\\
Hybrid edge var. & $6.6\%$ & $15.9\%$ & $2$ & n/a & $40$ & $10^{-3}$\\
  \rowcolor{Gray}
ARMANet & $\mathbf{2.0 \%}$ & $9.7\%$ & 1 & n/a &  $20$ & $10^{-3}$\\
GAT & $10.9\%$ & $20.8\%$ & n/a  &  $1$ & $40$ & $10^{-3}$\\
  \rowcolor{Gray}
GCAT & $8.0\%$ & $18.4$ &  $3$ & $1$ & $100$ & $10^{-3}$\\
Edge varying GAT & $7.1\%$ & $17.8\%$ & $2$ & $3$ & $100$ & $10^{-3}$\\
\Xhline{4\arrayrulewidth}
\end{tabular}
\label{tab:Fb_perf}\vskip-3mm
\end{table*}


Our goal is to see how the different architectures handle their degrees of freedom, while all having linear complexity. To make this comparison more insightful, we proceed with the following rationale. For the approaches in $i)-iv)$, we analyzed filter orders in the interval $K \in \{1, \ldots, 5\}$. This is the only handle we have on these filters to control the number of parameters and locality radius while keeping the same computational complexity. For the ARMANet in $v)$, we set the direct term order to $K = 0$ to observe only the effect of the rational part. Subsequently, for each Jacobi iteration value $K$, we analyzed rational orders in the interval $P \in \{1, \ldots, 5\}$ as for the former approaches. While this strategy helps us controlling the local radius, recall the ARMANet has a computational complexity slightly higher than the former four architectures. For the GAT in $vi)$, we analyzed different attention heads $R \in \{1, \ldots, 5\}$ such that the algorithm complexity matches those of the approaches $i)-iv)$. The number of attention heads is the only handle in the GAT network. Finally, for the GCAT in $vii)$ and the edge varying GAT in $viii)$, we fixed the attention heads to $R = 3$ and analyzed different filter orders $K \in \{1, \ldots, 5\}$. The latter allows comparing the impact of the local radius for the median attention head value. Recall, these architectures have again a slightly higher complexity for $K \ge 2$.

\medskip\noindent\textbf{Observations.} The results of this experiment are shown in Figure~\ref{fig:SourceLoc}. We make the following observations.

First, the attention-based approaches are characterized by a slow learning rate leading to a poor performance in $40$ epochs. This is reflected in the higher test error of the GAT and the edge varying GAT networks. However, this is not the case for the GCAT network. We attribute the latter reduced error to the superposition of the graph convolutional to attentions that GCAT explores --all convolutional approaches learn faster. 
On the contrary, the error increases further for the edge varying GAT since multiple attention strategies are adopted for all $k \in \{1, \ldots, K\}$ in \eqref{eqn_gat_edge_varying}. Therefore, our conclusion is the graph convolutional prior can be significantly helpful for attention mechanisms. We will see this consistent improvement in all our experiments.

Second, the edge varying GNN in \eqref{eqn_gnn_multiple_feature} achieves the lowest error, although having the largest number of parameters. The convolutional approaches parameterize well the edge varying filter; hence, highlighting the benefit of the graph convolution. ARMANet is the best among the latter characterized both by a lower mean error and standard deviation. This reduced error for ARMANet is not entirely surprising since rational functions have better interpolation and extrapolation properties than polynomial ones. It is, however, remarkable that the best result is obtained for a Jacobi iteration of $K = 1$. I.e., the parameter sharing imposed by ARMANet reaches a good local optimal even with a coarse approximation of the rational function. {Notice also the source localization task is not permutation equivariant, therefore, architectures that are not permutation equivariant (edge varying, node varying, hybrid edge varying, ARMA for low Jacobi orders) are expected to perform better.}

Third, for the node selection strategies in architectures $iii)$ and $iv)$, there is no clear difference between the degree and the diffusion centrality. For the node varying GNNs, the diffusion centrality offers a lower error both in the mean and deviation. In the hybrid edge varying GNNs [cf. \eqref{eq.nvGCNN1}], instead, the degree centrality achieves a lower error but pays in deviation. The spectral proxy centrality yields the worst performance.

Finally, we remark that we did not find any particular trend while changing the parameters of the different GNNs (e.g., order, attention head). A rough observation is that low order recursions are often sufficient to reach low errors. \vskip-2.5mm


\subsection{Source localization on Facebook sub-network}

In the second experiment, we considered a similar source localization on a real-world network comprising a $234-$used Facebook subgraph obtained as the largest connected component of the dataset in \cite{leskovec2012learning}. This graph has two well-defined connected communities of different size and the objective is to identify which of the two communities originated the diffusion. The performance of the different algorithms is averaged over $100$ Monte-Carlo iterations. {The remaining parameters for generating the synthetic data are similar as before.}

\medskip\noindent\textbf{Architecture parameters.} We compared the eight GNN architectures reported in the left-most column of Table~\ref{tab:Fb_perf}. For the node varying and the hybrid edge varying GNNs, the important nodes are again $10\%$ of all nodes selected based on diffusion centrality. The Jacobi number of iterations for the ARMANet is $K = 1$ while there is no direct term. 

{Overall, this problem is easy to solve if the GNN is hypertuned with enough width and depth. However, this strategy hinders the impact of the specific filter. To highlight the role of the latter, we considered minimal GNN architectures composed of one layer and two features. In turn, this allows understanding better how much the specific filter contributes to the performance.} We then grid-searched all parameters in Table~\ref{tab:Fb_perf} to reach a classification error of at most $2\%$. For the architectures that reach this criterion, we report the smallest parameters. For the architectures that do not reach this criterion, we report the minimum achieved error and the respective parameters. Our rationale is that the minimum parameters yield a lower complexity and show better the contribution of the filter type. They also lead to faster training; the opposite holds for the learning rate.

From Table~\ref{tab:Fb_perf}, we observe that only the edge varying GNN and the ARMANet reach the predefined error. Both architectures stress our observation that low order recursions ($K = 1$) are often sufficient. Nevertheless, this is not the case for all other architectures. {These observations suggest the edge varying GNN explores well its degrees of freedom and adapts well to the non-permutation equivariance of the task.} The ARMANet explores the best the convolutional prior; in accordance with the former results, the Jacobi implementation does not need to runt until convergence to achieve impressive results. We also conclude the convolutional prior helps to reduce the degrees of freedom of the EdgeNet but requires a deeper and/or wider network to achieve the predefined criterion. This is particularly seen in the GAT based architectures. The GCAT architecture, in here, explores the convolutional prior and reduces the error compared with the GAT. Finally, we remark for all approaches a substantially lower variance can be achieved by solely increasing the features.


\subsection{Authorship attribution}

{In this third experiment, we assess the performance of the different GNN architectures in an authorship attribution problem based on real data. The goal is to classify if a text excerpt belongs to a specific author or any other of the $20$ contemporary authors based on word adjacency networks (WANs) \cite{segarra15-wans}. A WAN is an author-specific directed graph whose nodes are function words without semantic meaning (e.g., prepositions, pronouns, conjunctions). The relative positioning of function words carries stylistic information about the author. To capture this information, we build a directed graph, where each node is a function word, and each weighted edge represents the average co-occurence of the corresponding function words, discounted by relative distance (i.e. if the two words are next to each other, the weight is higher than if the two words are further apart). We build this graph support only once, and before the training set. The signal on top of this graph is the frequency count for the function words in text excerpts of $1,000$ words. These are the graph signals that form the dataset that is used for training and testing. The WANs and the word frequency count serve as author signatures and allow learning representation patterns in their writing style. The task translates into a binary classification problem where one indicates the text excerpt is written by the author of interest and zero by any other author. A more detailed account on the creation of WANs can be found in \cite{segarra15-wans} and the used dataset is available in our code.}

The WANs of the respective authors have from $N = 190$ to $N = 210$ function word nodes. Following \cite{segarra15-wans}, we built single-author WANS for Jane Austen, Emily Bront\"{e}, and Edgar Allan Poe. For each author, we processed the texts to count the number of times each function word pair co-appears in a window of ten words. These co-appearances are imputed into an $N \times N$ matrix and normalized row-wise. The resulting matrix is used as the shift operator, which can also be interpreted as a Markov chain transition matrix. We considered a train-test split of $95\%-5\%$ of the available texts. Around $8.7\%$ of the training samples are used for validation. This division leads to: $(i)$ Austen: $1346$ training samples, $118$ validation samples, and $78$ testing samples; $(ii)$ Bront\"{e}: $1192$ training samples, $104$ validation samples, $68$ testing samples; $(iii)$ Poe: $740$ training samples, $64$ validation samples, $42$ testing samples. For each author, the sets are extended by a similar amount with texts from the other $20$ authors shared equally between them.

\medskip\noindent\textbf{Architecture parameters.} We considered again the eight GNN architectures of the former section shown in the leftmost column of Table~\ref{tab:Auth_perf}. Following the setup in \cite{gama2018convolutional}, all architectures comprise a graph neural layer of $F = 32$ features with ReLU nonlinearity followed by a fully connected layer. The baseline order for all filters is $K = 4$. For the ARMANet this is also the number of denominator parameters and the order of the direct term in \eqref{eqn_arma_jacobi_filter_1}; the number of the Jacobi iterations in \eqref{eqn_arma_jacobi_filter_2} is one. We want to show how much the rational part helps to improve the performance of the GCNN (which is the direct term in the ARMANet [cf. \eqref{eqn_arma_jacobi_filter_1}]). The important nodes for the node varying and the hybrid edge varying are $20$ ($\sim10\%$ of $N$) selected with degree centrality. The GAT, GCAT, and edge varying GAT have a single attention head to highlight the role of the convolutional and edge varying recursion over it. The loss function is the cross-entropy optimized over $25$ epochs with a learning rate of $0.005$. The performance is averaged over ten data splits.

Table~\ref{tab:Auth_perf} shows the results of this experiment. Overall, we see again the graph convolution is a solid prior to learning meaningful representations. This is particularly highlighted in the improved performance of the GCAT for Austen and Bront\"{e} compared with the GAT even with a single attention head. These observations also suggest the GAT and the edge varying GAT architectures require multi-head approaches to achieve comparable performance. An exception is the case of Poe. In this instance, multi-head attention is also needed for the GCAT. The (approximated) rational part of the ARMANet gives a consistent improvement of the GCNN. Hence, we recommend considering the additional parameterization of the ARMANet when implementing graph convolutional neural networks, since the increased number of parameters and implementation costs are minimal. Finally, we remark the hybrid edge varying GNN improves the accuracy of the node varying counterpart.\vskip-2.5mm


\begin{table}[!t]
\centering
\caption{Authorship Attribution Test Error. The results show the average classification test error and standard deviation on $10$ different training-test $95\%-5\%$ splits.}\vskip-.25cm
\begin{tabular}{l c c c}
\Xhline{4\arrayrulewidth}
Architecture & Austen & Bront\"{e} & Poe \\
\hline
  \rowcolor{Gray}
GCNN & $7.2 (\pm 2.0)\%$ & $12.9 (\pm 3.5)\%$  & $14.3 (\pm 6.4)\%$\\
Edge varying & $7.1 (\pm 2.2)\%$ & $13.1 (\pm 3.9)\%$ &  $\mathbf{10.7 (\pm 4.3)\%}$\\
  \rowcolor{Gray}
Node varying & $7.4 (\pm 2.1)\%$ & $14.6 (\pm 4.2)\%$  &  $11.7 (\pm 4.9)\%$ \\
Hybrid edge var.  & $\mathbf{6.9 (\pm 2.6)\%}$ & $14.0 (\pm 3.7)\%$ &  $11.7 (\pm 4.8)\%$\\
  \rowcolor{Gray}
ARMANet & $7.9 (\pm 2.3)\%$ & $\mathbf{11.6 (\pm 5.0)\%}$  &  ${10.9 (\pm 3.7)\%}$ \\
GAT  & $10.9 (\pm 4.6)\%$ & $22.1 (\pm 7.4)\%$ &  $12.6 (\pm 5.5)\%$\\
  \rowcolor{Gray}
GCAT & $8.2 (\pm 2.9)\%$ & $13.1 (\pm 3.5)\%$  &  $13.6 (\pm 5.8)\%$\\
Edge varying GAT & $14.5 (\pm 5.9)\%$ & $23.7 (\pm 9.0)\%$ & $18.1 (\pm 8.4)\%$\\
\hline
\Xhline{4\arrayrulewidth}
\end{tabular}
\label{tab:Auth_perf}\vskip-4mm
\end{table}



\subsection{Recommender Systems}

In this last experiment, we evaluate all former architectures for movie rating prediction in a subset of the MovieLens $100$K data set \cite{harper2015movielens}. The full data set comprises $U = 943$ users and $I = 1,582$ movies and $100$K out of $\sim\hskip-1mm 1,5$M potential ratings. We set the missing ratings to zero. From the incomplete $U\times I$ rating matrix, we consider two scenarios: a user-based and a movie-based. In a user-based scenario, we considered the $200$ users that have rated the most movies as the nodes of a graph whose edges represent Pearson similarities between any two users. Each of the $I = 1,582$ movies is treated as a different graph signal whose value at a node is the rating given to that movie by a user or zero if unrated. We are interested to predict the rating of a specific user $u$ with GNNs, which corresponds to completing the $u$th row of the $200 \times 1,5882$ sub-rating matrix. In a movie-based scenario, we considered the $200$ movies with the largest number of ratings as nodes of a graph whose edges represent Pearson similarities between any two movies. In this instance, there are $943$ graph signals: the ratings each user gives to all $200$ movies is one such graph signal. We are interested to predict the rating to a specific movie $i$ with GNNs, which corresponds to completing the $i$th column of the rating matrix. {We remark this task is permutation equivariant, therefore, we expect architectures holding this property to perform better.}


\begin{table}[!t]
\centering
\caption{Average RMSE on user graph.}\vskip-.25cm
\begin{tabular}{l c c c c c |c}
\Xhline{4\arrayrulewidth}
\!\!\!Archit./User-ID\!\! & $405$ & $655$ & $13$ & $450$ & $276$ & \!\!Average\!\!\! \\
\hline
  \rowcolor{Gray}
\!\!\!GCNN & $\mathbf{1.09}$ & $0.72$ & $1.18$ & $0.82$ & $0.66$ & $0.89$\!\!\!\\
\!\!\!Edge var. & $1.25$ & $0.74$ &  $1.34$ & $0.99$ &$0.70$ & $1.00$\!\!\! \\
  \rowcolor{Gray}
\!\!\!Node var. &$1.17$&$\mathbf{0.68}$&$1.19$&$0.83$&$0.67$ &$0.91$ \!\!\! \\
\!\!\!Hybrid edge var. 	& $\mathbf{1.10}$ & $0.72$ & $1.27$ & $0.80$ & $\mathbf{0.60}$ & $0.90$ \!\!\! \\
  \rowcolor{Gray}
\!\!\!ARMANet & $1.13$ & $\mathbf{0.69}$ & $1.24$ & $0.80$ & $0.65$ & $0.90$  \!\!\! \\
\!\!\!GAT  & $1.27$ & $0.74$ & $1.44$ & $0.92$ & $0.80$ & $1.03$ \!\!\! \\
  \rowcolor{Gray}
\!\!\!GCAT & $\mathbf{1.09}$ & $0.71$ & $\mathbf{1.12}$ & $\mathbf{0.77}$ & $0.65$ & $\mathbf{0.87}$ \!\!\! \\
\!\!\!Edge var. GAT & $1.19$ & $0.70$ & $1.31$ & $0.85$ & $0.75$ & $0.96$  \!\!\! \\
\hline
\Xhline{4\arrayrulewidth}
\end{tabular}
\label{tab:RecSys_userf}\vskip-4mm
\end{table}

\medskip\noindent\textbf{Architecture parameters.} We considered the same architectural settings as in the authorship attribution experiments to highlight consistent behaviors and differences. Following \cite{ruiz2019invariance}, we chose ten $90\%-10\%$ splits for training and test sets and pruned the graphs to keep only the top-$40$ most similar connections per node. The shift operator is again the adjacency matrix normalized by the maximum eigenvalue. The ADAM learning algorithm is run over $40$ epochs in batches of five and learning rate $5\times 10^{-3}$. We trained the networks on a smooth-$\ell_1$ loss and measure the accuracy through the root mean squared error (RMSE).

Tables~\ref{tab:RecSys_userf} and \ref{tab:RecSys_movie} show the results for the five users and five movies with the largest number of ratings, respectively. {The first thing to note is that GCAT consistently improves GAT. The latter further stresses that multi-head attentions are more needed in the GAT than in the GCAT.} Second, the edge varying GNN yields the worst performance because it is not a permutation equivariant architecture. {In fact, the node varying and the hybrid edge varying, which are  approaches in-between permutation equivariance and local detail, work much better.} This trend is observed also in the edge varying GAT results, suggesting that also the number of parameters in the edge varying is too high for this task.

{
\subsection{General Observations}\label{subsec.genobs}
 
 Altogether these experiments lead to four main observations.
 
\medskip\noindent\textbf{Edge varying GNN useful for non-permutation equivariant tasks.} The edge varying GNN [cf. \eqref{eqn_gnn_multiple_feature}] can perform well if the task is not permutation equivariant (e.g., source localization). We have observed that minimal architectures (i.e., lower number of features $F$, layers $L$ and filter order $K$) adapt easier to the task. This is because a minimal architecture has less degrees of freedom and can avoid overfitting. Contrarily, when the task is permutation equivariant the edge varying GNN will suffer and parameterizations matched to the task are needed.
 
\medskip\noindent\textbf{Convolution provides a strong parameterization.} In permutation equivariant tasks, we have seen GCNNs are a valid parameterization. They have shown potential also in tasks that are not permutation equivariant. However, contrarily to the edge varying GNN, GCNNs requite a wider architecture and with filters of higher order. We have also seen the ARMANet [cf. \eqref{eqn_arma}-\eqref{eqn_arma_jacobi_filter_2}] can improve the performance of the polynomial counterpart [cf \eqref{eqn_gcnn_matrix_notation}]. This is because for the rational frequency response ARMANet implements in each layer, thus requiring less parameters. However, surprisingly to us, we have observed that even with a few Jacobi iterations --ARMANets require at each layer to compute a matrix inverse, which we solve iteratively with the Jacobi method-- ARMANet can often achieve a better performance than the polynomial GCNN.
 
\medskip\noindent\textbf{Hybrid solutions offer a good parameterization.} While the edge varying GNN and the GCNN may outperform each other depending on the task, the node varying [cf. \eqref{eq.nvGCNN}] and the hybrid edge varying [cf. \eqref{eq.nvGCNN1}] has shown a good parameterization of the EdgeNet in all tasks. We have seen the hybrid edge varying to perform overall better than the node varying due to its edge-dependent parameters. However, we did not found any consistent difference in the node selection strategies. Our rationale is that the sampling strategy of the important nodes needs to be matched with the task at hand. 

\medskip\noindent\textbf{Graph convolutions improve attention.} The graph convolutional attention network [cf. \eqref{eqn_polyGat}] improves consistently over the GAT. This is because it generalizes the latter from a convolutional filter of order one to an arbitrary order. Working with the GCAT has shown to accelerate the learning procedure of the GAT and perform well in any task. In addition, the GCAT may result effective even with a single attention head. Instead, by generalizing the GAT to an edge varying GAT [cf. \eqref{eqn_gat_edge_varying}] we have not seen substantial improvement than that seen in the GCAT. This indicates that the attention mechanism may not be the best strategy to learn different shift operators in each layer.
}
\section{Conclusion}
\label{sec:conc}

This paper introduced EdgeNets: GNN architectures that allow each node to collect information from its direct neighbors and apply different weights to each of them. EdgeNets preserve the state-of-the-art implementation complexity and provide a single recursion that encompasses all state-of-the-art architectures. By showcasing how each solution is a particular instance of the EdgeNet, we provided guidelines to develop more expressive GNN architectures, yet without compromising the computational complexity. This paper, in specific, proposed eight GNN architectures that can be readily extended to scenarios containing multi-edge features.

The EdgeNet link showed a tight connection between the graph convolutional and graph attention mechanism, which have been so far treated as two separate approaches. We found the graph attention network learns the weight of a graph and then performs an order one convolution over this learned graph. Following this link, we introduced the concept of graph convolutional attention networks, which is an EdgeNet that jointly learns the edge weights and the parameters of a convolutional filter.

{We advocate the EdgeNet as a more formal way to build GNN solutions. However, further research is needed in three main directions. First, research should be done to explore the connection between the EdgeNets and receptive fields. This will lead to different parameterizations and architectures. Second, works needs to be done for assessing the capabilities of EdgeNets solutions to handle graph isomorphisms \cite{maron2019provably,morris2019weisfeiler}. Third, theoretical work is also needed to characterize how the stability of the EdgeNet to link perturbations \cite{gama2019stability,levie2019atransferability}.}
%
%
\vskip-4mm


\begin{table}[!t]
\centering
\caption{Average RMSE on movie graph.}\vskip-.25cm
\begin{tabular}{l c c c c c |c}
\Xhline{4\arrayrulewidth}
\!\!\!Archit./Movie-ID\!\! & $50$ & $258$ & $100$ & $181$ & $294$ & \!\!Average\!\!\! \\
\hline
  \rowcolor{Gray}
\!\!\!GCNN & $\mathbf{0.82}$ & $1.08$ & $\mathbf{0.95}$ & $0.86$ & $1.04$ & $0.95$\!\!\!\\
\!\!\!Edge var. & $0.93$ & $\mathbf{1.03}$ &  $1.00$ & $0.88$ &$1.24$ & $1.02$\!\!\! \\
  \rowcolor{Gray}
\!\!\!Node var. &$0.78$&$1.04$ & $1.00$ & $0.87$ & $\mathbf{1.00}$ &$\mathbf{0.94}$ \!\!\! \\
\!\!\!Hybrid edge var. 	& $0.75$ & $\mathbf{1.02}$ & $0.98$ & $\mathbf{0.82}$ & $1.08$ & $\mathbf{0.93}$\!\!\! \\
  \rowcolor{Gray}
\!\!\!ARMANet & $\mathbf{0.81}$ & $1.05$ & $1.02$ & $0.87$ & $1.09$ & $0.97$ \!\!\! \\
\!\!\!GAT  & $0.98$ & $1.24$ & $1.28$ & $1.00$ & $1.30$ & $1.16$ \!\!\! \\
  \rowcolor{Gray}
\!\!\!GCAT & $0.83$ & $1.06$ & $1.04$ & $\mathbf{0.83}$ & $1.05$ & $0.96$  \!\!\! \\
\!\!\!Edge var. GAT & $\mathbf{0.81}$ & $1.04$ & $1.01$ & $0.86$ & $1.07$ & $0.96$   \!\!\! \\
\hline
\Xhline{4\arrayrulewidth}
\end{tabular}
\label{tab:RecSys_movie}\vskip-4mm
\end{table}


\appendices
\label{sec:ginAPP}


\section{Proof of Proposition~\ref{prop:invariance}}\label{sec:proofP1}

Denote the respective graph shift operator matrices of the graphs $\ccalG$ and $\ccalG^\prime$ as $\bbS$ and $\bbS^\prime$. For $\bbP$ being a permutation matrix, $\bbS^\prime$ and $\bbx^\prime$ can be written as $\bbS^\prime = \bbP^\transp \bbS\bbP$ and $\bbx^\prime = \bbP^\transp\bbx$. Then, the output of the convolutional filter in \eqref{eqn_gcnn} applied to $\bbx^\prime$ is
\begin{equation}
\bbu^\prime = \sum_{k = 0}^Ka_k{\bbS^\prime}^k\bbx^\prime = \sum_{k = 0}^Ka_k\big(\bbP^\transp\bbS\bbP	\big)^k\bbP^\transp\bbx.
\end{equation}
By using the properties of the permutation matrix $\bbP^k = \bbP$ and $\bbP\bbP^\transp = \bbI_N$, the output $\bbu^\prime$ becomes
\begin{equation}\label{eq.permOut}
\bbu^\prime = \bbP^\transp\left(\sum_{k = 0}^Ka_k\bbS^k\bbx	\right) = \bbP^\transp\bbu
\end{equation}
which implies the filter output operating on the permuted graph $\ccalG^\prime$ with input $\bbx^\prime$ is simply the permutation of the convolutional filter in \eqref{eqn_gcnn} applied to $\bbx$. Subsequently, since the nonlinearities of each layer are pointwise they implicitly preserve permutation equivariance; hence, the output of a GCNN layer is a permuted likewise. These permutations will propagate in the cascade of the different layers yielding the final permuted output. \qed \vskip-4mm

\section{Proof of proposition~\ref{prop:EVResp}}\label{sec:AppPropEV}

To start, let $\check{\bbPhi}^{(1)} = \bbPhi^{(0)} + \bbPhi^{(1)}$ and $\check{\bbPhi}^{(k)} = \bbPhi^{(k)}$ for all $k = 2, \ldots, K$ be the transformed parameter matrices. Recall also that $\bbPhi^{(0)}$ is a diagonal matrix; thus, $\check{\bbPhi}^{(1)}$ shares the support with $\bbPhi^{(1)}$ and with $\bbS + \bbI_N$. Given the eigendecompostion of the transformed parameter matrices $\check{\bbPhi}^{(k)} = \bbV\bbLambda^{(k)}\bbV^{-1}$ for all $k = 1, \ldots, K$, the edge varying filter can be written in the graph spectral domain as
\begin{equation}\label{eq:EVFresp}
a(\bbLambda) = \sum_{k = 1}^K\bigg(\prod_{k^\prime = 1}^k\bbLambda^{(k^\prime)}	\bigg).
\end{equation}
Subsequently, recall that $\ccalJ$ is the index set defining the zero entries of $\bbS + \bbI_N$ and that $\bbC_\ccalJ \in \{0,1\}^{|\ccalI|\times N^2}$ is the selection matrix whose rows are those of $\bbI_{N^2}$ indexed by $\ccalJ$ [cf. \eqref{eqn.nulSpace}]. Then, the fixed support condition for $\check{\bbPhi}^{(k)}$ for all $k = 1, \ldots, K$ is
\begin{align}\label{eq.sparsSupp}
\begin{split}
\bbC_\ccalJ\text{vec}\left(\check{\bbPhi}^{(k)}\right) &= \! \bbC_\ccalJ\text{vec}\!\left(\!\bbV\bbLambda^{(k)}\bbV^{-1}\!\right) = \bbzero_{|\ccalJ|}.
\end{split}
\end{align}
%
Put differently, equation \eqref{eq.sparsSupp} expresses in a vector form the zero entries of $\check{\bbPhi}^{(k)}$ (hence, of $\bbPhi^{(0)}, \ldots, \bbPhi^{(K)}$) that match those of $\bbS + \bbI_N$. From the properties of the vectorization operation, \eqref{eq.sparsSupp} can be rewritten as
\begin{equation}\label{eq.evSpec}
\bbC_\ccalJ\text{vec}\!\left(\!\bbV\bbLambda^{(k)}\bbV^{-1}\!\right) \!=\! \bbC_\ccalJ\text{vec}\!\left(\!\bbV^{-1}\!*\!\bbV\right)\bblambda^{(k)}
\end{equation}
where $``*"$ denotes the Khatri-Rao product and $\bblambda^{(k)} = \diag(\bbLambda^{(k)})$ is the $N$-dimensional vector composed by the diagonal elements of $\bbLambda^{(k)}$. As it follows from \eqref{eq.sparsSupp}, \eqref{eq.evSpec} implies the vector $\bblambda^{(k)}$ lies in the null space of $\bbC_\ccalJ\text{vec}(\bbV^{-1}*\bbV)$, i.e., 
\begin{equation}\label{eq.basMat}
\bblambda^{(k)} \in \text{null}\big(\bbC_\ccalI\text{vec}(\bbV^{-1}*\bbV)\big).
\end{equation}

Let then $\bbB$ be a basis that spans spans this null space [cf. \eqref{eqn.nulSpace}]. The vector  $\bblambda^{(k)}$ can be expanded as 
\begin{equation}\label{eq:expLamk}
\bblambda^{(k)} = \bbB\bbmu^{(k)}
\end{equation}
$\bbmu^{(k)}$ is the vector containing the basis expansion parameters. Finally, by putting back together \eqref{eq.sparsSupp}-\eqref{eq:expLamk}, \eqref{eq:EVFresp} becomes
\begin{equation}\label{eq.EV_respfin}
a(\bbLambda) = \sum_{k=1}^{K}\prod_{k^\prime = 1}^k\diag\left(\bbB\bbmu^{(k^\prime)}\right).
\end{equation}

The $N \times b$ basis matrix $\bbB$ is a kernel that depends on the specific graph and in particular on the eigenvectors $\bbV$. The kernel dimension $b$ depends on the rank of $\bbB$ and thus, on the rank of the null space in \eqref{eq.basMat}. In practice it is often observed that $\rank(\bbB) = b \ll N$. \qed \vskip-4mm

%
%

\section{Diffusion centrality}~\label{sec:AppdiffCentr}

Let $\bbS$ be the shift operator used to represent the graph structure. We define the diffusion centrality (DC) $\delta_i$ of a node $i$ in $K$ shifts, as the $i$th entry of the vector
\begin{equation}\label{eq.diffCentr}
\bbdelta = \sum_{k = 0}^K\bbS^k\bbone_N.
\end{equation}
The DC describes how much each node influences the passing of information in the network for a finite time of hops. The DC vector $\bbdelta$ can also be seen as the convolution of the constant graph signal with a convolutional filter of the form in \eqref{eqn_gcnn} which has all unitary parameters. This definition of DC is more appropriate for choices of $\bbS$ being the adjacency matrix or normalizations of it. For $\bbS$ being the discrete Laplacian matrix, the DC is zero for all nodes since the constant vector is the eigenvector of the discrete Laplacian corresponding to the zero eigenvalue. The above DC definition is the particularization of the DC proposed in \cite{banerjee2013diffusion} for stochastic setting to the case where all nodes decide to take part in the signal diffusion. Both the DC in \eqref{eq.diffCentr} and the one from \cite{banerjee2013diffusion} are correlated to Katz-Bonacich centrality and eigenvector centrality.







%





\ifCLASSOPTIONcaptionsoff
  \newpage
\fi



\bibliographystyle{IEEEtran}
\bibliography{myIEEEabrv,edgeNetsBiblio}

\begin{thebibliography}{10}
\providecommand{\url}[1]{#1}
\csname url@samestyle\endcsname
\providecommand{\newblock}{\relax}
\providecommand{\bibinfo}[2]{#2}
\providecommand{\BIBentrySTDinterwordspacing}{\spaceskip=0pt\relax}
\providecommand{\BIBentryALTinterwordstretchfactor}{4}
\providecommand{\BIBentryALTinterwordspacing}{\spaceskip=\fontdimen2\font plus
\BIBentryALTinterwordstretchfactor\fontdimen3\font minus
  \fontdimen4\font\relax}
\providecommand{\BIBforeignlanguage}[2]{{%
\expandafter\ifx\csname l@#1\endcsname\relax
\typeout{** WARNING: IEEEtran.bst: No hyphenation pattern has been}%
\typeout{** loaded for the language `#1'. Using the pattern for}%
\typeout{** the default language instead.}%
\else
\language=\csname l@#1\endcsname
\fi
#2}}
\providecommand{\BIBdecl}{\relax}
\BIBdecl

\bibitem{isufi2019generalizing}
E.~Isufi, F.~Gama, and A.~Ribeiro, ``Generalizing graph convolutional neural
  networks with edge-variant recursions on graphs,'' in \emph{27th Eur. Signal
  Process. Conf.}\hskip 1em plus 0.5em minus 0.4em\relax A Coru\~{n}a, Spain:
  Eur. Assoc. Signal Process., 2-6 Sep. 2019.

\bibitem{tang2009relational}
L.~Tang and H.~Liu, ``Relational learning via latent social dimensions,'' in
  \emph{15th ACM SIGKDD Int. Conf. Knowledge Discovery and Data Mining}.\hskip
  1em plus 0.5em minus 0.4em\relax Paris, France: ACM, 28 June-1 July 2009, pp.
  817--826.

\bibitem{wale2008comparison}
N.~Wale, I.~A. Watson, and G.~Karypis, ``Comparison of descriptor spaces for
  chemical compound retrieval and classification,'' \emph{Knowledge and
  Information Systems}, vol.~14, no.~3, pp. 347--375, 2008.

\bibitem{bronstein2017geometric}
M.~M. Bronstein, J.~Bruna, Y.~LeCun, A.~Szlam, and P.~Vandergheynst,
  ``Geometric deep learning: Going beyond {Euclidean} data,'' \emph{{IEEE}
  Signal Process. Mag.}, vol.~34, no.~4, pp. 18--42, July 2017.

\bibitem{shuman2013emerging}
D.~I. Shuman, S.~K. Narang, P.~Frossard, A.~Ortega, and P.~Vandergheynst, ``The
  emerging field of signal processing on graphs: Extending high-dimensional
  data analysis to networks and other irregular domains,'' \emph{{IEEE} Signal
  Process. Mag.}, vol.~30, no.~3, pp. 83--98, May 2013.

\bibitem{scarselli2005graph}
F.~Scarselli, S.~L. Yong, M.~Gori, M.~Hagenbuchner, A.~C. Tso, and M.~Maggini,
  ``Graph neural networks for ranking web pages,'' in \emph{The 2005
  IEEE/WIC/ACM Int. Conf. Web Intelligence}.\hskip 1em plus 0.5em minus
  0.4em\relax Compiegne, France: IEEE, 19-22 Sep. 2005, pp. 1--7.

\bibitem{scarselli2008graph}
F.~Scarselli, M.~Gori, A.~C. Tsoi, M.~Hagenbuchner, and G.~Monfardini, ``The
  graph neural network model,'' \emph{{IEEE} Trans. Neural Netw.}, vol.~20,
  no.~1, pp. 61--80, Jan. 2009.

\bibitem{Gama20-GNNs}
F.~Gama, E.~Isufi, G.~Leus, and A.~Ribeiro, ``Graphs, convolutions, and neural
  networks: From graph filters to graph neural networks,'' \emph{{IEEE} Signal
  Process. Mag.}, vol.~37, no.~6, pp. 128--138, Nov. 2020.

\bibitem{bruna2013spectral}
J.~Bruna, W.~Zaremba, A.~Szlam, and Y.~LeCun, ``Spectral networks and deep
  locally connected networks on graphs,'' in \emph{2nd Int. Conf. Learning
  Representations}.\hskip 1em plus 0.5em minus 0.4em\relax Banff, AB: Assoc.
  Comput. Linguistics, 14-16 Apr. 2014, pp. 1--14.

\bibitem{defferrard2016convolutional}
M.~Defferrard, X.~Bresson, and P.~Vandergheynst, ``Convolutional neural
  networks on graphs with fast localized spectral filtering,'' in \emph{30th
  Conf. Neural Inform. Process. Syst.}\hskip 1em plus 0.5em minus 0.4em\relax
  Barcelona, Spain: Neural Inform. Process. Foundation, 5-10 Dec. 2016, pp.
  3844--3858.

\bibitem{gama2018convolutional}
F.~Gama, A.~G.~Marques, G.~Leus, and A.~Ribeiro, ``Convolutional neural network
  architectures for signals supported on graphs,'' \emph{{IEEE} Trans. Signal
  Process.}, vol.~67, no.~4, pp. 1034--1049, Feb. 2019.

\bibitem{du2017topology}
J.~Du, J.~Shi, S.~Kar, and J.~M.~F. Moura, ``On graph convolution for graph
  {CNNs},'' in \emph{2018 {IEEE} Data Sci. Workshop}.\hskip 1em plus 0.5em
  minus 0.4em\relax Lausanne, Switzerland: IEEE, 4-6 June 2018, pp. 239--243.

\bibitem{kipf2016semi}
T.~N. Kipf and M.~Welling, ``Semi-supervised classification with graph
  convolutional networks,'' in \emph{5th Int. Conf. Learning
  Representations}.\hskip 1em plus 0.5em minus 0.4em\relax Toulon, France:
  Assoc. Comput. Linguistics, 24-26 Apr. 2017, pp. 1--14.

\bibitem{xu2018powerful}
K.~Xu, W.~Hu, J.~Leskovec, and S.~Jegelka, ``How powerful are graph neural
  networks?'' in \emph{7th Int. Conf. Learning Representations}.\hskip 1em plus
  0.5em minus 0.4em\relax New Orleans, LA: Assoc. Comput. Linguistics, 6-9 May
  2019, pp. 1--17.

\bibitem{levie2017cayleynets}
R.~Levie, F.~Monti, X.~Bresson, and M.~M. Bronstein, ``{CayleyNets}: Graph
  convolutional neural networks with complex rational spectral filters,''
  \emph{{IEEE} Trans. Signal Process.}, vol.~67, no.~1, pp. 97--109, Jan. 2019.

\bibitem{simonovsky2017dynamic}
M.~Simonovsky and N.~Komodakis, ``Dynamic edge-conditioned filters in
  convolutional neural networks on graphs,'' in \emph{Conf. Comput. Vision and
  Pattern Recognition 2017}.\hskip 1em plus 0.5em minus 0.4em\relax Honolulu,
  HI: Comput. Vision Foundation, 21-26 July 2017, pp. 3693--3702.

\bibitem{monti2017geometric}
F.~Monti, D.~Boscaini, J.~Masci, E.~Rodol\`{a}, J.~Svoboda, and M.~M.
  Bronstein, ``Geometric deep learning on graphs and manifolds using mixture
  model {CNNs},'' in \emph{Conf. Comput. Vision and Pattern Recognition
  2017}.\hskip 1em plus 0.5em minus 0.4em\relax Honolulu, HI: Comput. Vision
  Foundation, 21-26 July 2017, pp. 3693--3702.

\bibitem{atwood2016diffusion}
J.~Atwood and D.~Towsley, ``Diffusion-convolutional neural networks,'' in
  \emph{30th Conf. Neural Inform. Process. Syst.}\hskip 1em plus 0.5em minus
  0.4em\relax Barcelona, Spain: Neural Inform. Process. Foundation, 5-10 Dec.
  2016.

\bibitem{velickovic2017graph}
P.~Veli{\v{c}}kovi{\'{c}}, G.~Cucurull, A.~Casanova, A.~Romero, P.~Li{\`{o}},
  and Y.~Bengio, ``Graph attention networks,'' in \emph{6th Int. Conf. Learning
  Representations}.\hskip 1em plus 0.5em minus 0.4em\relax Vancouver, BC:
  Assoc. Comput. Linguistics, 30 Apr.-3 May 2018, pp. 1--12.

\bibitem{wu2019comprehensive}
\BIBentryALTinterwordspacing
Z.~Wu, S.~Pan, F.~Chen, G.~Long, C.~Zhang, and P.~S. Yu, ``A comprehensive
  survey on graph neural networks,'' \emph{arXiv:1901.00596v3 [cs.LG]}, 8 Aug.
  2019. [Online]. Available: \url{http://arxiv.org/abs/1901.00596}
\BIBentrySTDinterwordspacing

\bibitem{zhou2018graph}
\BIBentryALTinterwordspacing
J.~Zhou, G.~Cui, Z.~Zhang, C.~Yang, Z.~Liu, L.~Wang, C.~Li, and M.~Sun, ``Graph
  neural networks: A review of methods and applications,''
  \emph{arXiv:1812.08434v4 [cs.LG]}, 10 July 2019. [Online]. Available:
  \url{http://arxiv.org/abs/1812.08434}
\BIBentrySTDinterwordspacing

\bibitem{zhang2018deep}
\BIBentryALTinterwordspacing
Z.~Zhang, P.~Cui, and W.~Zhu, ``Deep learning on graphs: A survey,''
  \emph{arXiv:1812.04202v1 [cs.LG]}, 11 Dec. 2018. [Online]. Available:
  \url{http://arxiv.org/abs/1812.04202}
\BIBentrySTDinterwordspacing

\bibitem{lee2018attention}
\BIBentryALTinterwordspacing
J.~B. Lee, R.~A. Rossi, S.~Kim, N.~K. Ahmed, and E.~Koh, ``Attention models in
  graphs: A survey,'' \emph{arXiv:1807.07984v1 [cs.AI]}, 20 July 2018.
  [Online]. Available: \url{http://arxiv.org/abs/1807.07984}
\BIBentrySTDinterwordspacing

\bibitem{taubiny2000geometric}
G.~Taubin, ``Geometric signal processing on polygonal meshes,'' in
  \emph{{EUROGRAPHICS '2000}}.\hskip 1em plus 0.5em minus 0.4em\relax
  Interlaken, Switzerland: The Eurographics Association, 21-25 Aug. 2000, pp.
  1--11.

\bibitem{shuman2011distributed}
D.~I. Shuman, P.~Vandergheynst, D.~Kressner, and P.~Frossard, ``Distributed
  signal processing via {Chebyshev} polynomial approximation,'' \emph{{IEEE}
  Trans. Signal, Inform. Process. Networks}, vol.~4, no.~4, pp. 736--751, Dec.
  2018.

\bibitem{sandryhaila2013discrete}
A.~Sandryhaila and J.~M.~F. Moura, ``Discrete signal processing on graphs,''
  \emph{{IEEE} Trans. Signal Process.}, vol.~61, no.~7, pp. 1644--1656, Apr.
  2013.

\bibitem{narang2013compact}
S.~K. Narang, ``Compact support biorthogonal wavelet filterbanks for arbitrary
  undirected graphs,'' \emph{{IEEE} Trans. Signal Process.}, vol.~61, no.~19,
  pp. 4673--4685, Oct. 2013.

\bibitem{teke2016extending}
O.~Teke and P.~P. Vaidyanathan, ``Extending classical multirate signal
  processing theory to graphs---{Part I}: Fundamentals,'' \emph{{IEEE} Trans.
  Signal Process.}, vol.~65, no.~2, pp. 409--422, Jan. 2017.

\bibitem{segarra2017optimal}
S.~Segarra, A.~G.~Marques, and A.~Ribeiro, ``Optimal graph-filter design and
  applications to distributed linear network operators,'' \emph{{IEEE} Trans.
  Signal Process.}, vol.~65, no.~15, pp. 4117--4131, Aug. 2017.

\bibitem{isufi2017autoregressive}
E.~Isufi, A.~Loukas, A.~Simonetto, and G.~Leus, ``Autoregressive moving average
  graph filtering,'' \emph{{IEEE} Trans. Signal Process.}, vol.~65, no.~2, pp.
  274--288, Jan. 2017.

\bibitem{you2020graph}
J.~You, J.~Leskovec, K.~He, and S.~Xie, ``Graph structure of neural networks,''
  in \emph{International Conference on Machine Learning}.\hskip 1em plus 0.5em
  minus 0.4em\relax PMLR, 2020, pp. 10\,881--10\,891.

\bibitem{bahdanau2014neural}
D.~Bahdanau, K.~Cho, and Y.~Bengio, ``Neural machine translation by jointly
  learning to align and translate,'' in \emph{3rd Int. Conf. Learning
  Representations}.\hskip 1em plus 0.5em minus 0.4em\relax San Diego, CA:
  Assoc. Comput. Linguistics, 7-9 May 2015, pp. 1--15.

\bibitem{vaswani2017attention}
A.~Vaswani, N.~Shazeer, N.~Parmar, J.~Uszkoreit, L.~Jones, A.~N. Gomez,
  {\L}.~Kaiser, and I.~Polosukhin, ``Attention is all you need,'' in \emph{31st
  Conf. Neural Inform. Process. Syst.}\hskip 1em plus 0.5em minus 0.4em\relax
  Long Beach, CA: Neural Inform. Process. Syst. Foundation, 4-9 Dec. 2017, pp.
  1--11.

\bibitem{coutino2018advances}
M.~Coutino, E.~Isufi, and G.~Leus, ``Advances in distributed graph filtering,''
  \emph{{IEEE} Trans. Signal Process.}, vol.~67, no.~9, pp. 2320--2333, May
  2019.

\bibitem{battaglia2018relational}
P.~W. Battaglia, J.~B. Hamrick, V.~Bapst, A.~Sanchez-Gonzalez, V.~Zambaldi,
  M.~Malinowski, A.~Tacchetti, D.~Raposo, A.~Santoro, R.~Faulkner
  \emph{et~al.}, ``Relational inductive biases, deep learning, and graph
  networks,'' \emph{arXiv preprint arXiv:1806.01261}, 2018.

\bibitem{gilmer2017neural}
J.~Gilmer, S.~S. Schoenholz, P.~F. Riley, O.~Vinyals, and G.~E. Dahl, ``Neural
  message passing for quantum chemistry,'' \emph{arXiv preprint
  arXiv:1704.01212}, 2017.

\bibitem{ortega2018graph}
A.~Ortega, P.~Frossard, J.~Kova{\v{c}}evi{\'c}, J.~M. Moura, and
  P.~Vandergheynst, ``Graph signal processing: Overview, challenges, and
  applications,'' \emph{Proceedings of the IEEE}, vol. 106, no.~5, pp.
  808--828, 2018.

\bibitem{yan2018spatial}
S.~Yan, Y.~Xiong, and D.~Lin, ``Spatial temporal graph convolutional networks
  for skeleton-based action recognition,'' in \emph{Proceedings of the AAAI
  Conference on Artificial Intelligence}, vol.~32, no.~1, 2018.

\bibitem{hamilton2017inductive}
W.~Hamilton, Z.~Ying, and J.~Leskovec, ``Inductive representation learning on
  large graphs,'' in \emph{Advances in neural information processing systems},
  2017, pp. 1024--1034.

\bibitem{ioannidis2019recurrent}
V.~N. Ioannidis, A.~G.~Marques, and G.~B. Giannakis, ``A recurrent graph neural
  network for multi-relational data,'' in \emph{44th {IEEE} Int. Conf. Acoust.,
  Speech and Signal Process.}\hskip 1em plus 0.5em minus 0.4em\relax Brighton,
  UK: IEEE, 12-17 May 2019, pp. 8157--8161.

\bibitem{maron2018invariant}
H.~Maron, H.~Ben-Hamu, N.~Shamir, and Y.~Lipman, ``Invariant and equivariant
  graph networks,'' \emph{arXiv preprint arXiv:1812.09902}, 2018.

\bibitem{levie2019transferability}
R.~Levie, E.~Isufi, and G.~Kutyniok, ``On the transferability of spectral graph
  filters,'' in \emph{13th Int. Conf. Sampling Theory Applications}.\hskip 1em
  plus 0.5em minus 0.4em\relax Bordeaux, France: IEEE, 8-12 July 2019, pp.
  1--5.

\bibitem{gama2019stability}
F.~Gama, J.~Bruna, and A.~Ribeiro, ``Stability properties of graph neural
  networks,'' \emph{{IEEE} Trans. Signal Process.}, vol.~68, pp. 5680--5695, 25
  Sep. 2020.

\bibitem{gama2018convolutionalNV}
F.~Gama, G.~Leus, A.~G.~Marques, and A.~Ribeiro, ``Convolutional neural
  networks via node-varying graph filters,'' in \emph{2018 {IEEE} Data Sci.
  Workshop}.\hskip 1em plus 0.5em minus 0.4em\relax Lausanne, Switzerland:
  IEEE, 4-6 June 2018, pp. 220--224.

\bibitem{sandryhaila2014discrete}
A.~Sandyhaila and J.~M.~F. Moura, ``Discrete signal processing on graphs:
  Frequency analysis,'' \emph{{IEEE} Trans. Signal Process.}, vol.~62, no.~12,
  pp. 3042--3054, June 2014.

\bibitem{liu2008hadamard}
S.~Liu and G.~Trenkler, ``Hadamard, khatri-rao, kronecker and other matrix
  products,'' \emph{International Journal of Information and Systems Sciences},
  vol.~4, no.~1, pp. 160--177, 2008.

\bibitem{heckert2003nist}
N.~A. Heckert and J.~J. Filliben, ``Nist/sematech e-handbook of statistical
  methods; chapter 1: Exploratory data analysis,'' 2003.

\bibitem{trefethen2019approximation}
L.~N. Trefethen, \emph{Approximation Theory and Approximation Practice,
  Extended Edition}.\hskip 1em plus 0.5em minus 0.4em\relax SIAM, 2019.

\bibitem{liu2018filter}
J.~Liu, E.~Isufi, and G.~Leus, ``Filter design for autoregressive moving
  average graph filters,'' \emph{{IEEE} Trans. Signal, Inform. Process.
  Networks}, vol.~5, no.~1, pp. 47--60, March 2019.

\bibitem{isufi2017autoregressivest}
E.~Isufi, A.~Loukas, and G.~Leus, ``Autoregressive moving average graph
  filtering: A stable distributed implementation,'' in \emph{IEEE International
  Conference on Acoustics, Speech and Signal Processing (ICASSP)}, no.
  EPFL-CONF-223825, 2017.

\bibitem{bianchi2019graph}
\BIBentryALTinterwordspacing
F.~M. Bianchi, D.~Grattarola, C.~Alippi, and L.~Livi, ``Graph neural networks
  with convolutional {ARMA} filters,'' \emph{arXiv:1901.01343v5 [cs.LG]}, 24
  Oct. 2019. [Online]. Available: \url{http://arxiv.org/abs/1901.01345}
\BIBentrySTDinterwordspacing

\bibitem{wijesinghe2019dfnets}
W.~A.~S. Wijesinghe and Q.~Wang, ``Dfnets: Spectral cnns for graphs with
  feedback-looped filters,'' in \emph{Advances in Neural Information Processing
  Systems}, 2019, pp. 6007--6018.

\bibitem{ruiz2020gated}
L.~Ruiz, F.~Gama, and A.~Ribeiro, ``Gated graph recurrent neural networks,''
  \emph{{IEEE} Trans. Signal Process.}, vol.~68, pp. 6303--6318, 26 Oct. 2020.

\bibitem{gregor2010learning}
K.~Gregor and Y.~LeCun, ``Learning fast approximations of sparse coding,'' in
  \emph{Proceedings of the 27th international conference on international
  conference on machine learning}, 2010, pp. 399--406.

\bibitem{kingma2014adam}
D.~P. Kingma and J.~L. Ba, ``{ADAM}: A method for stochastic optimization,'' in
  \emph{3rd Int. Conf. Learning Representations}.\hskip 1em plus 0.5em minus
  0.4em\relax San Diego, CA: Assoc. Comput. Linguistics, 7-9 May 2015, pp.
  1--15.

\bibitem{anis2016efficient}
A.~Anis, A.~Gadde, and A.~Ortega, ``Efficient sampling set selection for
  bandlimited graph signals using graph spectral proxies,'' \emph{{IEEE} Trans.
  Signal Process.}, vol.~64, no.~14, pp. 3775--3789, July 2016.

\bibitem{leskovec2012learning}
J.~McAuley and J.~Leskovec, ``Learning to discover social circles in {Ego}
  networks,'' in \emph{26th Conf. Neural Inform. Process. Syst.}\hskip 1em plus
  0.5em minus 0.4em\relax Stateline, TX: Neural Inform. Process. Foundation,
  3-8 Dec. 2012.

\bibitem{segarra15-wans}
S.~Segarra, M.~Eisen, and A.~Ribeiro, ``Authorship attribution through function
  word adjacency networks,'' \emph{{IEEE} Trans. Signal Process.}, vol.~63,
  no.~20, pp. 5464--5478, Oct. 2015.

\bibitem{harper2015movielens}
F.~M. Harper and J.~A. Konstan, ``The movielens datasets: History and
  context,'' \emph{Acm transactions on interactive intelligent systems (tiis)},
  vol.~5, no.~4, pp. 1--19, 2015.

\bibitem{ruiz2019invariance}
L.~Ruiz, F.~Gama, A.~G.~Marques, and A.~Ribeiro, ``Invariance-preserving
  localized activation functions for graph neural networks,'' \emph{{IEEE}
  Trans. Signal Process.}, vol.~68, pp. 127--141, 25 Nov. 2019.

\bibitem{maron2019provably}
H.~Maron, H.~Ben-Hamu, H.~Serviansky, and Y.~Lipman, ``Provably powerful graph
  networks,'' in \emph{Advances in neural information processing systems},
  2019, pp. 2156--2167.

\bibitem{morris2019weisfeiler}
C.~Morris, M.~Ritzert, M.~Fey, W.~L. Hamilton, J.~E. Lenssen, G.~Rattan, and
  M.~Grohe, ``Weisfeiler and leman go neural: Higher-order graph neural
  networks,'' in \emph{Proceedings of the AAAI Conference on Artificial
  Intelligence}, vol.~33, 2019, pp. 4602--4609.

\bibitem{levie2019atransferability}
R.~Levie, W.~Huang, L.~Bucci, M.~M. Bronstein, and G.~Kutyniok,
  ``Transferability of spectral graph convolutional neural networks,''
  \emph{arXiv preprint arXiv:1907.12972}, 2019.

\bibitem{banerjee2013diffusion}
A.~Banerjee, A.~G. Chandrasekhar, E.~Duflo, and M.~O. Jackson, ``The diffusion
  of microfinance,'' \emph{Science}, vol. 341, no. 6144, pp. 1\,236\,498
  (1--7), July 2013.

\end{thebibliography}
\end{document}